%% file: main.tex
\documentclass[10pt]{article} % For LaTeX2e
\usepackage[preprint]{tmlr}
\input{math_commands.tex}

\usepackage{hyperref}
\usepackage{url}

\usepackage{graphicx}
\usepackage{subfigure}
\usepackage{booktabs}
\usepackage{resizegather}
\usepackage{wrapfig}
\usepackage{multirow}

\graphicspath{{./images/}}

\title{Multi-Grid Tensorized Fourier Neural Operator for High-Resolution PDEs}

\author{\name Jean Kossaifi \email jkossaifi@nvidia.com \\
      \addr NVIDIA
      \AND
      \name Nikola Kovachki \email nkovachki@nvidia.com \\
      \addr NVIDIA
      \AND
      \name Kamyar Azizzadenesheli \email kamyara@nvidia.com\\
      \addr NVIDIA \\
      \name Anima Anandkumar \email aanandkumar@nvidia.com\\
      \addr NVIDIA \\
      }

\def\month{MM}  % Insert correct month for camera-ready version
\def\year{YYYY} % Insert correct year for camera-ready version
\def\openreview{\url{https://openreview.net/forum?id=XXXX}} % Insert correct link to OpenReview for camera-ready version

\begin{document}

\maketitle

\input{sections/0-Abstract}
\input{sections/1-Introduction}
\input{sections/Background}
\input{sections/Method}
\input{sections/experiments-setting}
\input{sections/experiments-results}
\input{sections/conclusions}

\bibliography{references}
\bibliographystyle{tmlr}

% \appendix
% \section{Appendix}
% \input{sections/appendix}

\end{document}

%% file: math_commands.tex
\usepackage{amsmath,amsfonts,bm}
\usepackage{xspace}

\newcommand{\FNO}{\textnormal{\textsl{FNO}}\xspace}
\newcommand{\TOP}{\textnormal{\textsl{TFNO}}\xspace}
\newcommand{\TNO}{\textnormal{\textsl{TFNO}}\xspace}
\newcommand{\PDE}{\textnormal{\textsl{PDE}}\xspace}

\newcommand{\alg}{\textnormal{\textsl{MG-TFNO}}\xspace}

\newcommand{\F}{\mathcal F}
\newcommand{\Real}{\mathbb{R}}
\newcommand{\A}{\mathcal A}
\newcommand{\U}{\mathcal{U}}
\newcommand{\G}{\mathcal{G}}

\newcommand{\TT}{\mathbb{T}}
\newcommand{\tensor}[1]{{\mathbf{#1}}}

\renewcommand{\matrix}[1]{\ensuremath{\mathbf{#1}}}
\newcommand{\kmax}{\alpha}
\newcommand{\Id}{\operatorname{Id}}

\newcommand{\Natural}{\mathbb{N}}
\def\eqref#1{equation~\ref{#1}}
\def\1{\bm{1}}

\DeclareMathAlphabet{\mathsfit}{\encodingdefault}{\sfdefault}{m}{sl}
\SetMathAlphabet{\mathsfit}{bold}{\encodingdefault}{\sfdefault}{bx}{n}

\newcommand{\Cov}{\mathrm{Cov}}

%% file: sections/0-Abstract.tex
\begin{abstract}
Memory complexity and data scarcity have so far prohibited learning solution operators of partial differential equations (\PDE) at high resolutions. 
We address these limitations by introducing a new data efficient and highly parallelizable operator learning approach with reduced memory requirement and better generalization, called multi-grid tensorized neural operator (\alg). 
\alg scales to large resolutions by leveraging local and global structures of full-scale, real-world phenomena, through a decomposition of both the input domain and the operator's parameter space. 
Our contributions are threefold: i) we enable parallelization over input samples with a novel multi-grid-based domain decomposition,
ii) we represent the parameters of the model in a high-order latent subspace of the Fourier domain, through a global tensor factorization, resulting in an extreme reduction in the number of parameters and improved generalization, 
and iii) we propose architectural improvements to the backbone \FNO. Our approach can be used in any operator learning setting.
We demonstrate superior performance on the turbulent Navier-Stokes equations where we achieve less than half the error with over 150$\times$ compression. The tensorization combined with the domain decomposition, yields over 150$\times$ reduction in the number of parameters and $7\times$ reduction in the domain size without losses in accuracy,  while slightly enabling parallelism.
\end{abstract}

%% file: sections/1-Introduction.tex
 \section{Introduction}
Real-world scientific computing problems often time require repeatedly solving large-scale and high-resolution partial differential equations (\PDE{}s). For instance, in weather forecasts, large systems of differential equations are solved to forecast the future state of the weather. Due to internal inherent and aleatoric uncertainties, multiple repeated runs are carried out by meteorologists every day to quantify prediction uncertainties. Conventional \PDE solvers constitute the mainstream approach used to tackle such computational problems. However, these methods are known to be slow and memory-intensive. They require an immense amount of computing power, are unable to learn and adapt based on observed data, and oftentimes require sophisticated tuning~\citep{slingo2011uncertainty,leutbecher2008ensemble,blanusa2022role}.

Neural operators are a new class of models that aim at tackling these challenging problems~\citep{Graph}. They are maps between function spaces whose trained models emulate the solution operators of  \PDE{}s~\citep{universal}. In the context of \PDE{}s, these deep learning models are orders of magnitude faster than conventional solvers, can easily learn from data, can incorporate physically relevant information, and recently enabled solving problems deemed to be unsolvable with the current state of available \PDE{} methodologies \citep{liu2022learning,PINO}. Among neural operator models, Fourier neural operators (\FNO{}s), in particular, have seen successful application in scientific computing for the task of learning the solution operator to \PDE{}s as well as in computer vision for classification, in-painting, and segmentation \citep{FNO, kovachki2021onuniversal, guibas2021adaptive}. By leveraging spectral theory, \FNO{}s have successfully advanced frontiers in weather forecasts, carbon storage, and seismology~\citep{pathak2022fourcastnet,CCS,yang2021seismic}. 

While \FNO{}s have shown tremendous speed-up over classical numerical methods, their efficacy can be limited due to the rapid growth in memory needed to represent complex operators. In the worst case, large memory complexity is required and, in fact, is unavoidable due to the need for resolving fine-scale features globally.
However, many real-world problems, possess a local structure not currently exploited by neural operator methods. For instance, consider a weather forecast where predictions for the next hour are heavily dependent on the weather conditions in local regions and minimally on global weather conditions. Incorporating and learning this local structure of the underlying \PDE{}s is the key to overcoming the curse of memory complexity.

% This growth may become a bottleneck in their application to high-resolution physical simulations such as climate or materials modeling. In general, despite significant speed-up and     flexibility,  prior works on neural operators suffer from similar memory complexity issues as conventional solvers do on high-resolution problems.

\begin{figure}
    \centering
    \includegraphics[width=0.7\linewidth]{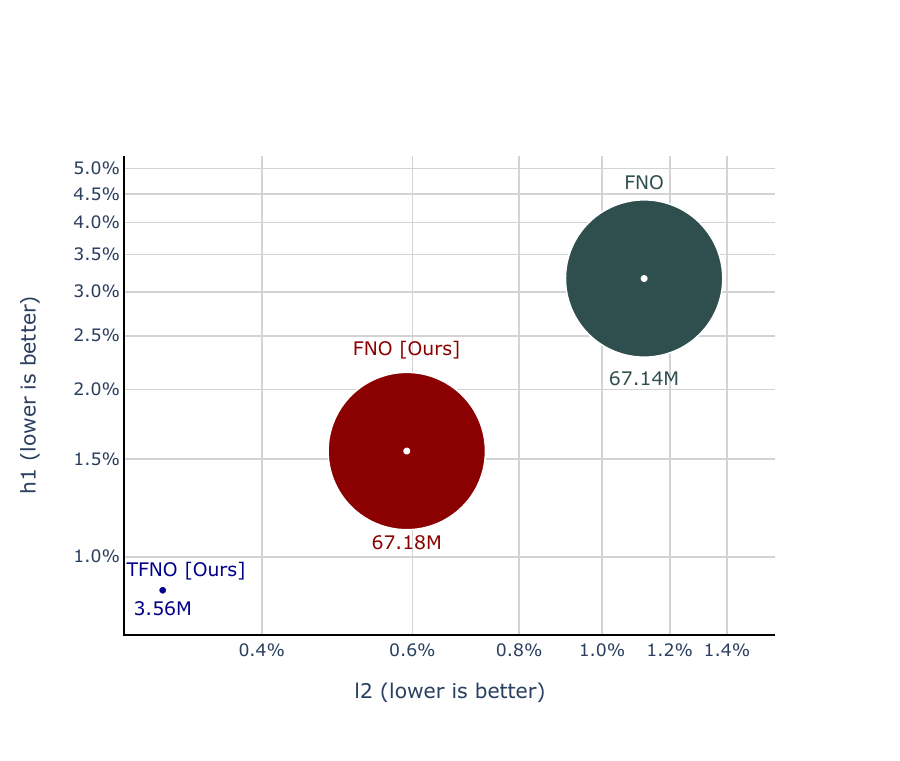}
    \caption{\textbf{Comparison of the performance on the relative $L^2$ and $H^1$ test errors (lower is better) on a log-scale} of our approach, compared with both our improved backbone (\emph{FNO}) and the original FNO, on Navier-Stokes. Our approach enables large compression for both input and parameter, while outperforming regular \FNO.}
    \label{fig:comparison_best}
\end{figure}

\begin{figure*}
    \centering
    \includegraphics[width=1\textwidth]{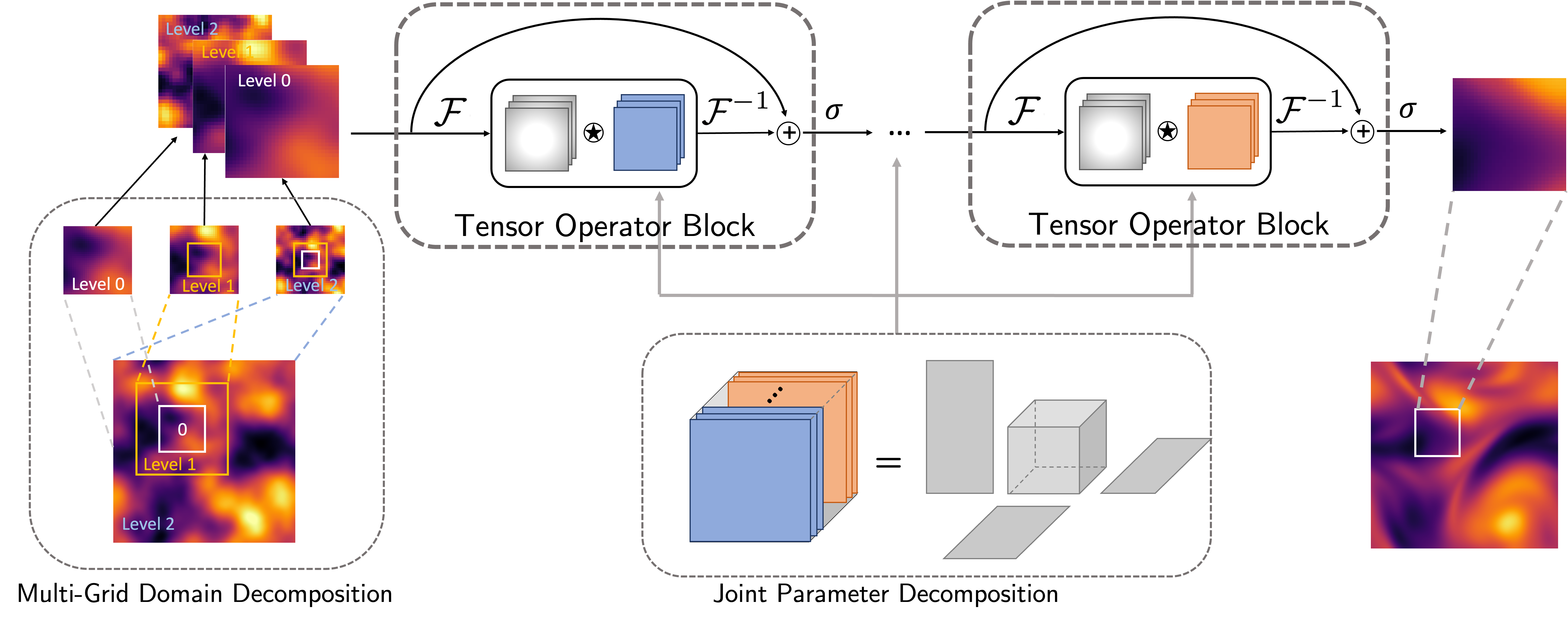}
    \caption{\textbf{Overview of our approach}. First (left), a multi-grid approach is used to create coarse to fine inputs that capture high-resolution details in a local region while still encoding global context. The resulting regions are fed to a tensorized Fourier operator (middle), the parameters of which are jointly represented in a single latent space via a low-rank tensor factorization (here, a Tucker form). Here $\F$ denotes Fourier transform. Finally, the outputs (right) are stitched back together to form the full result. Smoothness in the output is ensured via the choice of the loss function.}
    \label{fig:main_figure}
\end{figure*}

In this work, we propose a new, scalable neural operator that addresses these issues by leveraging the structure in both the domain space and the parameter space, Figure~\ref{fig:main_figure}. Specifically, we introduce the multi-grid tensor operator (\alg), a model that exploits locality in physical space by a novel multi-grid domain decomposition approach to compress the input domain size by up to $7\times$ while leveraging the global interactions of the model parameters to compress them by over $100\times$ without any loss of accuracy.% (see Table~\ref{tab:overall_results}).

\textbf{In the input space}, to predict the solution in any region of the domain, \alg decomposes the input domain into small local regions to which hierarchical levels of global information are added in a multi-grid fashion. Since a local prediction depends most strongly on its immediate spatial surroundings, the farther field information is downsampled to lower resolutions, progressively, based on its distance from the region of interest. Thus, \alg allows parallelization over the input domain as it relies on high-resolution data only locally and coarse-resolution data globally. Due to its state-of-the-art performance on \PDE{} problems and efficient FFT-based implementation, we use the \FNO{} as the backbone architecture for our method. It is worth noting that the multi-grid approach is readily amendable to neural network settings and, moreover, any other neural operator architecture can be used in place of \FNO as a backbone. 

\textbf{In the parameter space}, we exploit the spatiotemporal structure of the underlying \PDE solution operator by parameterizing the convolutional weights within the Fourier domain with a low-rank tensor factorization. Specifically, we impose a coupling between all the weights in the Fourier space by jointly parameterizing them with a single tensor, learned in a factorized form such as Tucker or Canonical-Polyadic~\citep{kolda2009tensor}. This coupling allows us to limit the number of parameters in the model without limiting its expressivity. On the contrary, this low-rank regularization on the model mitigates over-fitting and improves generalization. Intuitively, our method can be thought of as a fully-learned implicit scheme capable of converging in a small, fixed number of iterations. Due to the global nature of the integral kernel transform, the \FNO avoids the Courant–Friedrichs–Lewy (CFL) condition plaguing explicit schemes, allowing convergence in only a few steps~\citep{courant1928uber}. Our weight coupling ensures maximum communication between the steps, mitigating possible redundancies in the learned kernels and reducing the complexity of the optimization landscape.

\textbf{In summary, we make the following contributions:} %\vspace{-5pt}
\begin{itemize} %[leftmargin=9pt]
    \item \textbf{We propose architectural improvements to the backbone} which we validated through thorough ablations. %Using those, \alg largely outperforms \FNO with only a fraction of the parameters.
    \item \textbf{We propose} \alg, a novel neural operator parameterized in the spectral domain by a single low-rank factorized tensor, allowing its size to grow linearly with the size of the problem.%, Section~\ref{sec:top}
    \item \textbf{Our tensor operator achieves better performance with a fraction of the parameters}: we outperform \FNO on solving the turbulent Navier Stokes equations with more than $400\times$ weight compression ratio, Figure~\ref{fig:compression}.
    \item \textbf{Our method overfits less and does better in the low-data regime}. In particular, it outperforms \FNO with less than half the training samples, Figure~\ref{fig:low-data-regime}.
    \item  \textbf{We introduce a novel multi-grid domain decomposition approach}, a technique which allows the operator to predict the output only on local portions of the domain, thus reducing the memory usage by an order of magnitude with no performance degradation.%Section~\ref{fig:compression}.
    \item \textbf{Combining tensorization with multi-grid domain decomposition leads to \alg}, which is more efficient in terms of task performance, computation, and memory. \alg achieves $2.5\times$ lower error with $10\times$ model weight compression, and $1.8\times$ domain compression. %, Table~\ref{tab:overall_results}.
    \item \textbf{A unified codebase} to run all configurations and variations of \FNO and \alg will be released, along with the Navier-Stokes data used in this paper.
\end{itemize}

%% file: sections/Background.tex
\section{Background}
Here, we review related works and introduce the background necessary to explain our approach.

% \paragraph{Neural Operators.}
Many physical phenomena are governed by \PDE{}s and a wide range of scientific and engineering computation problems are based on solving these equations. In recent years, a new perspective to \PDE{}s dictates to formulate these problems as machine learning problems where solutions to \PDE{}s are learned. Prior works mainly focused on using neural networks to train for the solution map of \PDE{}s~\citep{guo2016convolutional,Zabaras,Adler2017,bhatnagar2019prediction,gupta2021multiwavelet}. The use of neural networks in the prior works limits them to a fixed grid and narrows their applicability to \PDE{}s where maps between function spaces are desirable. Multiple attempts have been made to address this limitation. For example mesh free methods are proposed that locally output mesh-free solution~\citep{lu2019deeponet,esmaeilzadeh2020meshfreeflownet}, but they are still limited to fixed input gird.  

A new deep learning paradigm, neural operators, are proposed as maps between function spaces~\citep{Graph, universal}. They are discretization invariants maps. The input functions to neural operators can be presented in any discretization, mesh, resolution, or basis. The output functions can be evaluated at any point in the domain. Variants of neural operators deploy a variety of Nystr\"om approximation to develop new neural operator architecture. Among these, multi-pole neural operators~\citep{Multipole} utilize the multi-pole approach to develop computationally efficient neural operator architecture. Inspired by the spectral method, Fourier-based neural operators show significant applicability in practical applications~\citep{FNO,yang2021seismic,CCS,rahman2022generative}, and the architectures have been used in neural networks for vision and text tasks~\citep{guibas2021adaptive,dao2022monarch}. Principle component analysis and u-shaped methods are also considered~\citep{bhattacharya2020model,liu2022learning,rahman2022u,yang2022inversion}. It is also shown that neural operators can solely be trained using \PDE{}s, resulting in physics-informed neural operators, opening new venues for hybrid data and equation methods~\citep{PINO} to tackle problems in scientific computing. 

Decomposing the domain in smaller subdomains is at the core of many methods in computational sciences\citep{Chan1994Domain} and extensively developed in deep learning~\citep{dosovitskiy2020image}. Prior deep learning methods on neural networks propose to decompose the input finite dimension vector to multiple patches, accomplish local operations, and aggregate the result of such process in the global sense~\citep{dosovitskiy2020image,guibas2021adaptive}. Such methods do not decompose the output domain and directly predict the entire output vector. In contrast, \alg works on function spaces, and not only decomposes the input domain, but also decomposes the domain of the output functions, and separately predicts the output at each subdomain. 

As we move beyond learning from simple structures to solving increasingly complex problems, the data we manipulate becomes more structured. To efficiently manipulate these structures, we need to go beyond matrix algebra and leverage the spatiotemporal structure. For all purposes of this paper, tensors are multi-dimensional arrays and generalize the concept of matrices to more than 2 modes (dimensions). For instance, RGB images are encoded as third-order (three-dimensional) tensors, videos are $4^{\text{th}}$ order tensors and so on and so forth. Tensor methods generalize linear algebraic methods to these higher-order structures. They have been very successful in various applications in computer vision, signal processing, data mining and machine learning~\citep{tensor2021panagakis,janzamin2019spectral,sidiropoulos2017tensor,papalexakis2016tensors}.

Using tensor decomposition~\cite{kolda2009tensor}, previous works have been able to compress and improve deep networks for vision tasks.
Either a weight matrix is tensorized and factorized~\citet{novikov2015tensorizing}, or tensor decomposition is directly to the convolutional kernels before fine-tuning to recover-for lost accuracy, which also allows for an efficient reparametrization of the network~\citep{lebedev2015speeding,yong2016compression,Gusak_2019_ICCV}. There is a tight link between efficient convolutional blocks and tensor factorization and factorized higher-order structures~\citep{kossaifi2020factorized}. Similar strategies have been applied to multi-task learning~\citep{bulat2020incremental} and NLP~\citep{papadopoulos2022efficient,cordonnier2020multi}.
Of all these prior works, none has been applied to neural operator. In this work, we propose the first application of tensor compression to learning operators and propose a Tensor OPerator (\TOP).

%% file: sections/Method.tex
\section{Methodology}
Here, we briefly review operator learning as well as the Fourier Neural Operator, on which we build to introduce our proposed Tensor OPerator (\TOP) as well as the Multi-Grid Domain Decomposition, which together form our proposed \alg.

\subsection{Operator Learning}
Let $\A:=\lbrace a:D_\A\rightarrow \Real^{d_\A}\rbrace$ and $\U:=\lbrace u:D_\U\rightarrow \Real^{d_\U}\rbrace$ denote two input and output function spaces respectively. Each function $a$, in the input function space $\A$, is a map from a bounded, open set $D_\A\subset\Real^{d}$ to the $d_\A$-dimensional Euclidean space. Any function in the output function space $\U$ is a map from a bounded open set $D_\U\subset\Real^{d}$ to the $d_\U$-dimensional Euclidean space. In this work we consider the case $D=D_\A=D_\U\subset \Real^{d}$. 
% Moreover, for any $p\in \N$, let $[p]$ denote the set $\lbrace1,\ldots,p\rbrace$. 

We aim to learn an operator $\G: \A \to \U$ which is a mapping between the two function spaces. In particular, given a dataset of $N$  points $\lbrace(a_j,u_j)\rbrace_{j=1}^N$, where the pair \((a_j,u_j)\) are functions satisfying \(\G(a_j)=u_j\), we build an approximation of the operator $\G$ . As a backbone operator learning model, we use neural operators as they are consistent and universal learners in function spaces. For an overview of theory and implementation, we refer the reader to \cite{universal}. We specifically use the \FNO and give details in the forthcoming section \citep{FNO}. 

\subsection{Notation}
\label{sec:app_notation}
We summarize the notation used throughout the paper in Table~\ref{tab:notation}.
\input{tables/notation}

\subsection{Fourier Neural Operators}

For simplicity, we will work on the \(d\)-dimensional unit torus \(\TT^d\) and first describe a single, pre-activation \FNO layer mapping \(\Real^m\)-valued functions to \(\Real^n\)-valued functions. Such a layer constitutes the mapping \(\G : L^2(\TT^d;\Real^m) \to L^2(\TT^d;\Real^n)\) defined as
\begin{equation}
\label{eq:FNO}
\G(v) = \F^{-1} \big ( \F(\kappa) \cdot \F(v) \big ), \qquad \forall \: v \in  L^2(\TT^d;\Real^m)
\end{equation}
where \(\kappa \in L^2 (\TT^d; \Real^{n \times m})\) is a function constituting the layer parameters and \(\F, \F^{-1}\) are the Fourier transform and its inverse respectively. The Fourier transform of the function \(\kappa\) is parameterized directly by some fixed number of Fourier nodes denoted  \(\kmax \in \Natural\).

To implement \eqref{eq:FNO}, \(\F, \F^{-1}\) are replaced by the discrete fast Fourier transforms \(\hat{\F}, \hat{\F}^{-1}\).
Let \(\hat{v} \in \Real^{s_1 \times \cdots \times s_d \times m}\) denote the evaluation of the function \(v\) on a uniform grid 
discretizing \(\TT^d\) with \(s_j \in \Natural\) points in each direction. We replace \(\F(\kappa)\) with a weight tensor \(\tensor{T} \in \Cov^{s_1 \times \cdots \times s_d \times n \times m}\) consisting of the Fourier modes of \(\kappa\) which are parameters to be learned. To ensure that \(\kappa\) is parameterized as a \(\Real^{n \times m}\)-valued function with a fixed, maximum amount of wavenumbers \(\kmax < \frac{1}{G2} \min \{s_1,\cdots,s_d\}\) that is independent of the discretization of \(\TT^d\), we leave as learnable parameters only the first \(\kmax\) entries of \(\tensor{T}\) in each direction and enforce that \(\tensor{T}\) have conjugate symmetry. In particular, we parameterize half the corners of the \(d\)-dimensional hyperrectangle with \(2^{d-1}\) hypercubes with length size \(\kmax\). That is, \(\tensor{T}\) is made up of the free-parameter tensors \(\tensor{\tilde{T}}_1,\cdots,\tensor{\tilde{T}}_{2^{d-1}} \in \Cov^{\kmax \times \cdots \times \kmax \times n \times m}\) situated in half of the corners of \(\tensor{T}\). Each corner diagonally opposite of a tensor \(\tensor{\tilde{T}}_j\) is assigned the conjugate transpose values of \(\tensor{\tilde{T}_j}\). All other values of \(\tensor{T}\) are set to zero. This is illustrated in the middle-top part of Figure~\ref{fig:main_figure} for the case \(d=2\) with \(\tensor{\tilde{T}}_1\) and \(\tensor{\tilde{T}}_2\). We will use the notation \(\tensor{T}(k, \cdots) = \tensor{\tilde{T}}_k\) for any \(k \in [2^{d-1}]\).
The discrete version of \eqref{eq:FNO} then becomes the mapping \(\hat{\G} : \Real^{s_1 \times \cdots \times s_d \times m} \to \Real^{s_1 \times \cdots \times s_d \times n}\) defined as 
\begin{equation}
    \label{eq:FNOdiscrete}
    \hat{\G}(\hat{v}) = \hat{\F}^{-1} \big ( \tensor{T} \cdot \hat{\F} (\hat{v}) \big ), \qquad \forall \: \hat{v} \in \Real^{s_1 \times \cdots \times s_d \times m}
\end{equation}
where the \(\cdot\) operation is simply the matrix multiplication contraction along the last dimension. Specifically, we have

\begin{equation}
\label{eq:contraction}
\big (\tensor{T} \cdot \hat{\F} (\hat{v}) \big ) (l_1,\dots,l_d,j) = 
\sum_{i=1}^m \tensor{T}(l_1,\dots,l_d,j,i) \big ( \hat{\F}(\hat{v}) \big )(l_1,\dots,l_d,i).
\end{equation}

From \eqref{eq:FNOdiscrete}, a full FNO layer is build by adding a point-wise linear action to \(\hat{v}\), a bias term, and applying a non-linear activation.
In particular, from an input \(\hat{v} \in \Real^{s_1 \times \cdots \times s_d \times m}\), the output \(\hat{q} \in \Real^{s_1 \times \cdots \times s_d \times n}\) is given as 
\[\hat{q}(l_1,\cdots,l_d,:) = \sigma \big ( \tensor{Q} \hat{v}(l_1, \cdots, l_d, :) + \hat{\G}(\hat{v}) + b \big ) \]
with \(\sigma : \Real \to \Real\) a fixed, non-linear activation, and \(b \in \Real^n\), \(\tensor{Q} \in \Real^{n \times m}\), \(\tensor{\tilde{T}}_1,\cdots,\tensor{\tilde{T}}_{2^{d-1}} \in \Cov^{\kmax \times \cdots \times \kmax \times n \times m}\) are the learnable parameters of the layer. The full FNO model consists of \(L \in \Natural\) such layers each with weight tensors \(\tensor{T}_1, \cdots,\tensor{{T}}_L\) that have learnable parameters \(\tensor{\tilde{T}}^{(l)}_k = \tensor{T}_l(k, \cdots)\) for any \(l \in [L]\) and \(k \in [2^{d-1}]\). In the case \(n=m\) for all layers, we introduce the joint parameter tensor \(\tensor{W} \in \Cov^{\kmax \times \cdots \times \kmax \times n \times n \times 2^{d-1} L}\) so that
\[
\tensor{W} \left(\dots,  2^{d-1}(l-1) + k + 1 \right) = \tensor{\tilde{T}}^{(l)}_k.
\]

A perusal of the above discussion reveals that there are \((2^d \kmax^d + 1)m n +  n\) total parameters in each FNO layer. Note that, since \(m\) and \(n\) constitute the respective input and output channels of the layer, the number of parameters can quickly explode due to the exponential scaling factor \(2^d \kmax^d\) if many wavenumbers are kept. Preserving a large number of modes could be crucial for applications where the spectral decay of the input or output functions is slow such as in image processing or the modeling of multi-scale physics. In the following section, we describe a tensorization method that is able to mitigate this growth without sacrificing approximation power. 

\subsection{Architectural improvements}
\label{sec:architecture}

Our proposed approach uses \FNO as a backbone. To improve its performance, we first study various aspects of the Fourier Neural Architecture and perform thorough ablation to validate each aspect. In particular, we propose improvements to the base architecture that improve performance. 

\begin{figure}
    \centering
    \subfigure[\textbf{Original FNO.}]{%
        \label{fig:original_arch}
        \includegraphics[height=7cm]{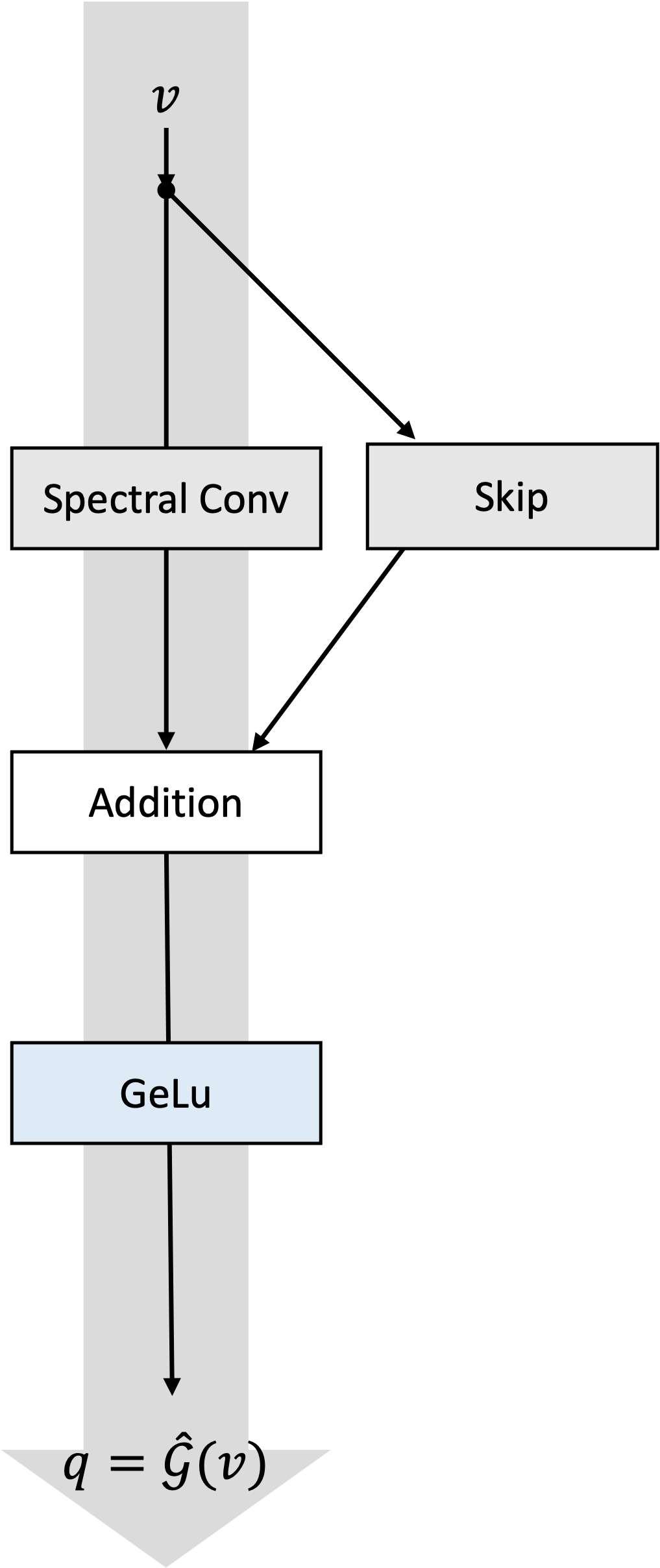}
    }%
    \hfill
    \subfigure[\textbf{Double Skip.}]{
        \label{fig:arch_double_skip}
        \includegraphics[height=7cm]{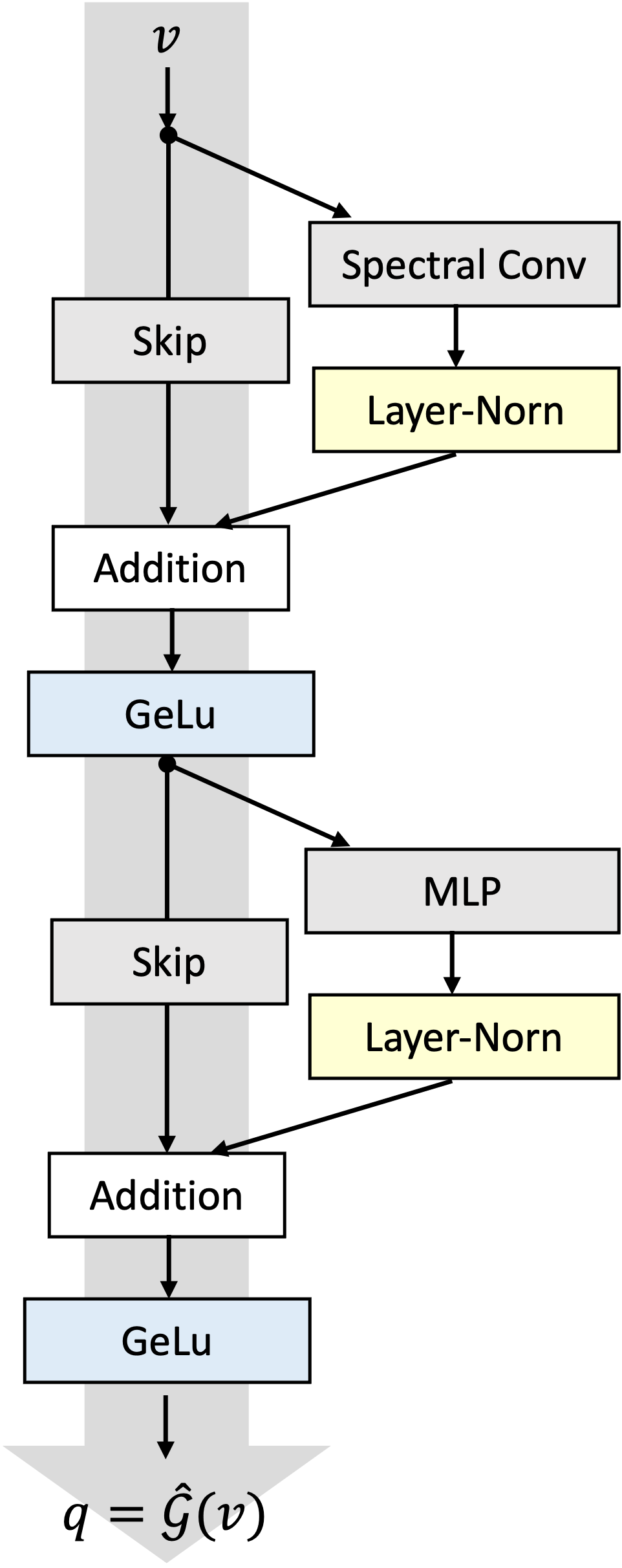}
    }
    \hfill
    \subfigure[\textbf{Improved backbone - preactivation.}]{
        \label{fig:improved_arch_preact}
        \includegraphics[height=7cm]{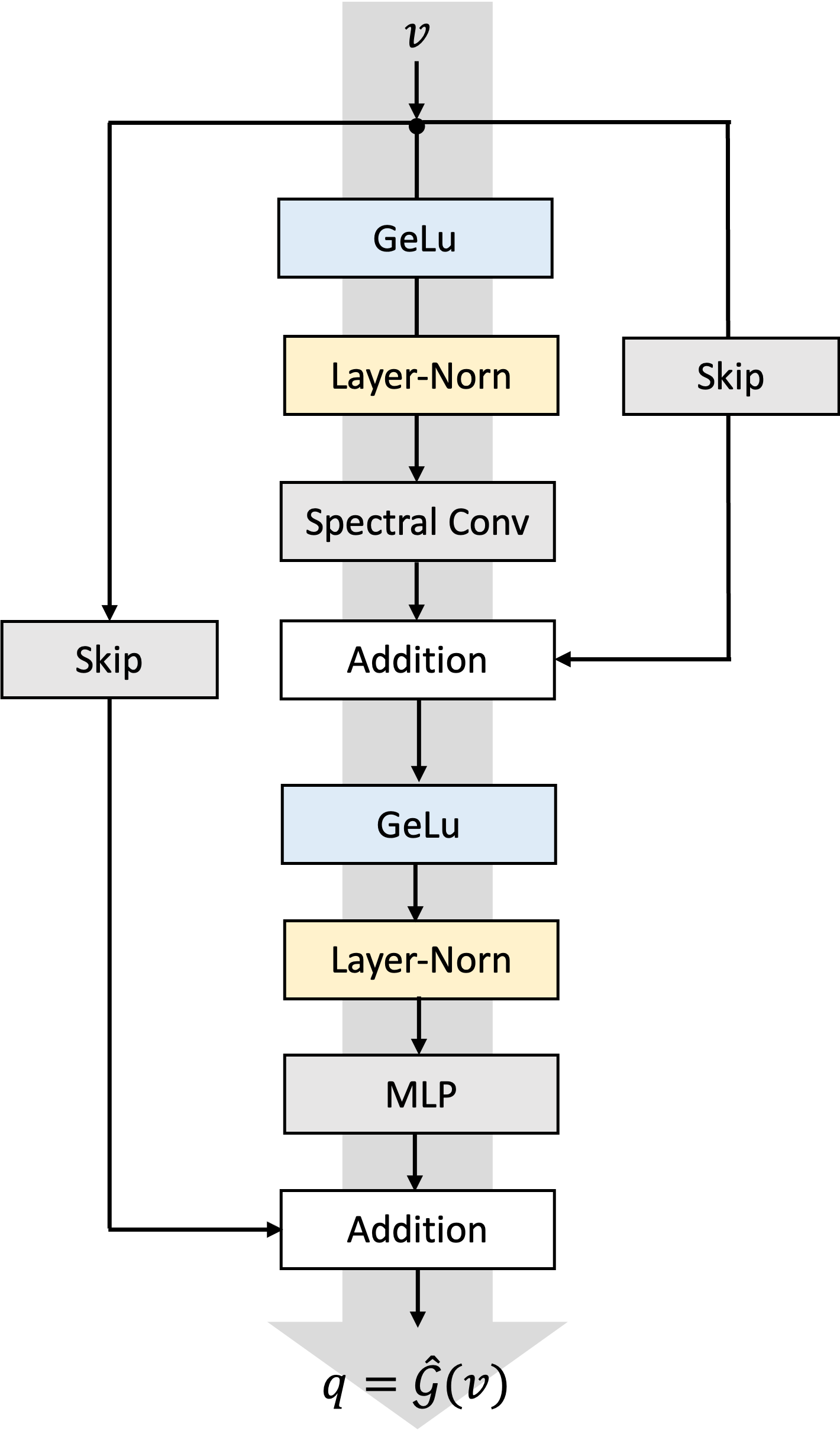}
    }   
    \hfill
    \subfigure[\textbf{Improved backbone.}]{
        \label{fig:improved_arch}
        \includegraphics[height=7cm]{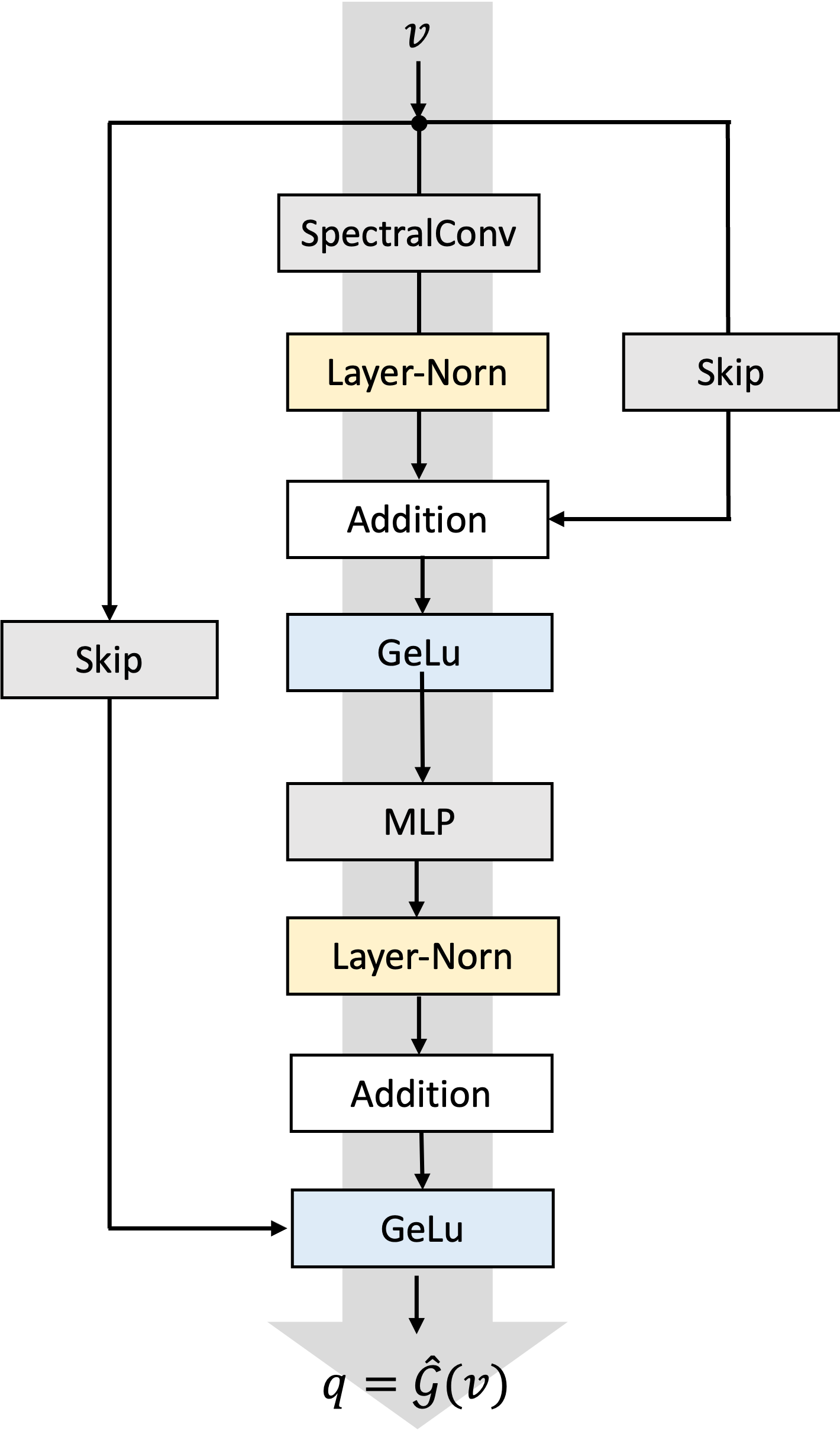}
    }   
    \caption{\textbf{Original FNO and Improved Backbone Architecture.}
    The original FNO architecture~\citep{FNO} is composed of simply a Spectral Convolution, with a (linear) skip connection to recover high-frequency information and handle non-periodic inputs (\ref{fig:original_arch}). We improve the architecture as detailed in section~\ref{sec:architecture}. In particular, we have a version with a double (sequential) skip connection (\ref{fig:arch_double_skip}), while our best architecture uses nested skip connections, and can be made both with and without preactivation (subfigures~\ref{fig:improved_arch_preact} and~\ref{fig:improved_arch}, respectively). The latter, subfigure~\ref{fig:improved_arch}, is our best architecture.
    }
    \label{fig:architecture}
\end{figure}

\paragraph{Normalization in neural operators} While normalization techniques, such as Batch-Normalization~\cite{batchnorm}, have proven very successful in training neural networks, additional consideration must be given when applying those to neural operators in order to preserve its properties, notably discretization invariance. Specifically, it cannot depend on the spatial variables and therefore has to be either a global or a function-wise normalization. We investigate several configurations using instance normalization~\cite{instancenrom} and layer-normalization~\cite{layernorm}, in conjunction with the use-of preactivation~\cite{he2016identity}.

\paragraph{Channel mixing} \FNO relies on a global convolution realized in the spectral domain. Inspired by previous works, e.g.~\cite{guibas2021adaptive}, we propose adding an MLP in the \emph{original} space, after each Spectral convolution. In practice, we found that two-layer bottleneck MLP works well, e.g. we decrease the co-dimension by half in the first linear layer before restoring it in the second one.

\paragraph{Boundary conditions} Fourier neural operators circumvent the limitation of traditional Fourier methods to inputs with periodic boundaries only. This is achieved through a local linear transformation added to the spectral convolution. This can be seen as a linear skip connection. We investigate replacing these with an identity skip-connection and a soft-gated skip-connection~\cite{bulat2020toward}. 

We also investigate the impact of domain-padding, found by~\citet{FNO} to improve results, especially for non-periodic inputs, and padding for the multi-grid decomposition.

We represent in Figure.~\ref{fig:architecture} the original FNO architecture~\citep{FNO}, subfigure~\ref{fig:original_arch}, the improved version with double (sequential) skip connections (subfigure~\ref{fig:arch_double_skip}) and our best architecture, both with and without preactivation (subfigures~\ref{fig:improved_arch_preact} and~\ref{fig:improved_arch}, respectively).
% \paragraph{Empirical results}
% We compare the original FNO to our \TOP with the improved backbone, with a width of 64, and using all the modes in the spectral domain. For these experiments, we use a mini-batch size of $16$ and no domain padding, which we found to give the best performance when the inputs are already periodic, which is the case for Navier-Stokes. We find that those improvements lead to significantly improved performance for all models. %, as can be seen in Table~\ref{tab:best}. 
% Our improved \FNO outperforms largely the original version~\cite{FNO}, and our \TOP, using this improved backbone, achieves improved performance over the backbone with over 180 times compression.

\subsection{Tensor Fourier Neural Operators}
\label{sec:top}
In the previous section, we introduced a unified formulation of FNO where the whole operator is parametrized by a single parameter tensor $\tensor{W}$. This enables us to introduce the tensor operator, which parameterizes efficiently $\tensor{W}$ with a low-rank, tensor factorization. We introduce the method for the case of a Tucker decomposition, for its flexibility. Other decompositions, such as Canonical Polyadic, can be readily integrated. This joint parametrization has several advantages: i) it applies a low-rank constraint on the entire tensor $\tensor{W}$, thus regularizing the model. These advantages translate into i) a huge reduction in the number of parameters, ii) better generalization and an operator less prone to overfitting. We show superior performance for low-compression ratios (up to $200\times$) and very little performance degradation when largely compressing ($>450\times$) the model, iii) better performance in a low-data regime.

In practice, we express $\tensor{W}$ in a low-rank factorized form, e.g. Tucker or CP. In the case of a Tucker factorization with rank $(R_1, \cdots, R_d, R_L, R_I, R_O)$, where $R_L$ controls the rank across layers, $R_I = R_O$ control the rank across the input and output co-dimension, respectively, and $R_1, \cdots, R_d$ control the rank across the dimensions of the operator:

\begin{gather}
    \tensor{W} = 
        \sum_{r_1 = 1}^{R_1}
        \cdots
        \sum_{r_d = 1}^{R_d}
        \sum_{r_i = 1}^{R_I}
        \sum_{r_o = 1}^{R_O}
        \sum_{r_l = 1}^{R_L}
        \tensor{G}(r_1, \cdots, r_d, r_i, r_o, r_l) \cdot
        \matrix{U^{(1)}}(:, r_1) \cdot \,\,
        \cdots \,\,\cdot
        \matrix{U^{(d)}}(:, r_d) \cdot
        \matrix{U^{(I)}}(:, r_i) \cdot
        \cdot \matrix{U^{(O)}}(:, r_o) \cdot 
        \matrix{U^{(L)}}(:, r_l).
\end{gather}

Here, $\tensor{G}$ is the core of size $R_L \times R_I \times R_O \times R_1 \times \cdots \times R_d$ and $\matrix{U^{(L)}}, \matrix{U^{(I)}}, \matrix{U^{(O)}}, \matrix{U^{(1)}}, \cdots, \matrix{U^{(d)}}$ are factor matrices of size $(R_L \times L), (R_I \times I), (R_O \times O), (R_1 \times \kmax), \cdots, (R_d \times \kmax)$, respectively. 

Note that the mode (dimension) corresponding to the co-dimension can be left uncompressed by setting $R_L = L$ and $\matrix{U^{(L)}} = \Id$. This leads to layerwise compression. Also note that having a rank of $1$ along any of the modes would mean that the slices along that mode differ only by a (multiplicative) scaling parameter. Also note that during the forward pass, we can pass $\tensor{T}$ directly in factorized form to each layer by selecting the corresponding rows in $\matrix{U^{(L)}}$. While the contraction in equation~\ref{eq:contraction} can be done using the reconstructed tensor, it can also be done directly by contracting $ \hat{\F}(\hat{v})$ with the factors of the decomposition. For small, adequately chosen ranks, this can result in computational speedups.

\begin{wrapfigure}[12]{rH}{0.35\columnwidth}
\begin{minipage}{0.35\textwidth}
% \begin{figure}
    \centering
    \includegraphics[width=\columnwidth]{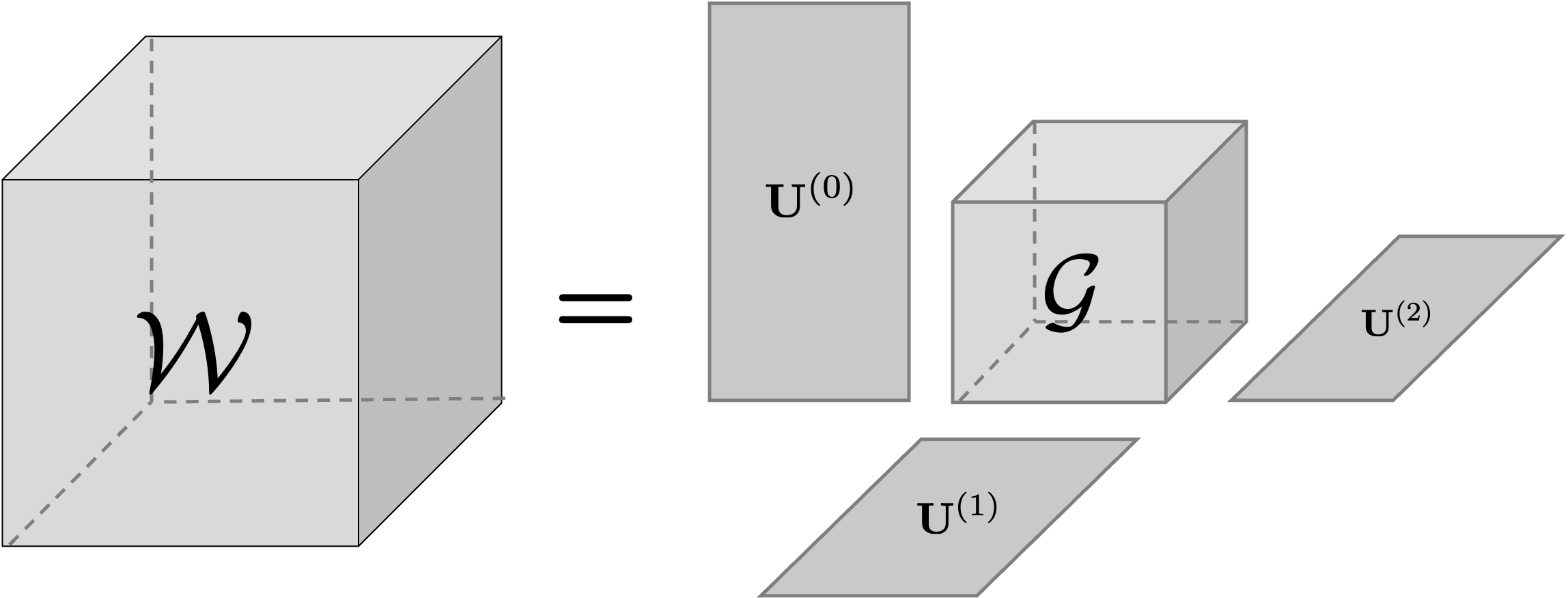}
    \caption{\textbf{Illustration of a Tucker decomposition.} For clarity , we show $\tensor{W}$ as a $3^{\text{rd}}$-order tensor weight.}
    \label{fig:tucker}
% \end{figure}
\end{minipage}
\end{wrapfigure}

A visualization of the Tucker decomposition of a third-order tensor can be seen in Figure~\ref{fig:tucker}). Note that we can rewrite the entire weight parameter for this Tucker case, equivalently, using the more compact n-mode product as: 
\begin{align*}
    \tensor{W} =
    \tensor{G} \times_1 \matrix{U^{(1)}}
    \cdots
    \times_d  \matrix{U^{(d)}}
    \times_{d+1} \matrix{U^{(I)}}
    \times_{d+2} \matrix{U^{(O)}}
    \times_{d+3} \matrix{U^{(L)}}
\end{align*}

We can efficiently perform an iFFT after contraction with the tensorized kernel. For any layer $l$, the $(j_1,j_2)$ coordinate of the matrix-valued convolution function $\kappa(x)$ is as follows,
\begin{align*}
        [\kappa_l(x)]{j1,j_2} =&
        % \left(\frac{1}{2\pi}\right)^d
        \sum_{i_1 = 1}^{m_1}
        \cdots
        \sum_{i_d = 1}^{m_d}
        \sum_{r_l = 1}^{R_L}
        \sum_{r_i = 1}^{R_I}
        \sum_{r_o = 1}^{R_O}
        \sum_{r_1 = 1}^{R_1}
        \cdots
        \sum_{r_d = 1}^{R_d}
        \tensor{G}(r_1, \cdots, r_d, r_i, r_o, r_l)\cdot \\
        &\matrix{U^{(1)}}(i_1, r_1)
        \cdots
        \matrix{U^{(d)}}(i_d, r_d) \cdot
        \matrix{U^{(I)}}(j_1, r_i) \cdot
        \matrix{U^{(O)}}(j_2, r_o) \cdot
        \matrix{U^{(L)}}(l, r_l)   \cdot
        \exp(2\pi\sum_{k = 1}^{d}ix_ki_k)
\end{align*}

This joint factorization along the entire operator allows us to leverage redundancies both locally and across the entire operator. This leads to a large reduction in the memory footprint, with only a fraction of the parameter. It also acts as a low-rank regularizer on the operator, facilitating training. Finally, through global parametrization, we introduce skip connections that allow gradients to flow through the latent parametrization to all the layers jointly, leading to better optimization. 

%%% FIGURE Multi-grid 
\begin{figure*}[htb]
    \subfigure[\textbf{Predicting with padded regions.} Local region in the input is padded and used to predict the corresponding region in the output. ]{%
        \label{fig:patching_only}
        \includegraphics[width=0.55\linewidth]{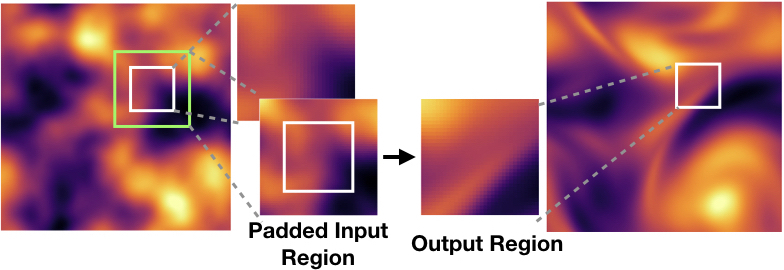}
    }%
    \hfill
    \subfigure[\textbf{MG-Domain Decomposition.} Progressively larger spatial regions are added to a local region by subsampling.]{
        \label{fig:multigrid_patching}
        \includegraphics[width=0.35\linewidth]{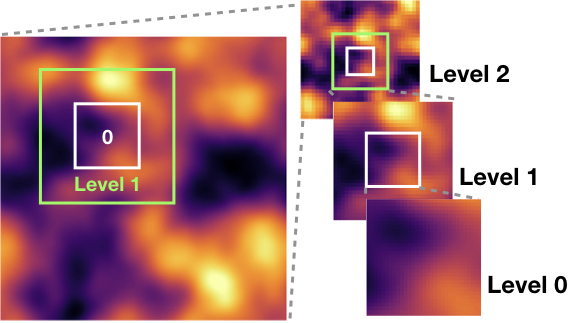}
    }   
    \caption{\textbf{Domain decomposition in space (\ref{fig:patching_only}) and our Multi-Grid based approach.
    (\ref{fig:multigrid_patching})}. White squares represent the region of interest while yellow squares the larger embeddings.}
\end{figure*}

Importantly, this formulation is general and works with any tensor factorization. For instance, we also explore a Canonical-Polyadic decomposition (CP) which can be seen as a special case of Tucker with a super-diagonal core. In that case, we set a single rank $R$ and express the weights as a weighted sum of $R$ rank-1 tensors. Concretely:
\begin{align}
    \tensor{W} = 
        \sum_{r = 1}^{R}
        \lambda_r &
        \matrix{U^{(1)}}(:, r) \cdot \,\,
        \cdots \,\,\cdot
        \matrix{U^{(d)}}(:, r) \cdot\\ \nonumber
        & \matrix{U^{(I)}}(:, r) \cdot
        \matrix{U^{(O)}}(:, r) \cdot 
        \matrix{U^{(L)}}(:, r).
\end{align}
where $\matrix{U^{(L)}}, \matrix{U^{(I)}}, \matrix{U^{(O)}}, \matrix{U^{(1)}}, \cdots, \matrix{U^{(d)}}$ are factor matrices of size $(R \times L), (R \times I), (R \times O), (R \times \kmax), \cdots, (R \times \kmax)$, respectively and $\bf{\lambda} \in \mathbb{R}^R$. Note that the CP, contrarily to the Tucker, has a single rank parameter, shared between all the dimensions. This means that to maintain the number of parameters the same, $R$ needs to be very high, which leads to memory issues. This makes CP more suitable for large compression ratios, and indeed, we found it leads to better performance at high-compression / very low-rank.  In this paper, we also explore the tensor-train decomposition~\cite{oseledets2011tensor}. A rank-$(1, R_1, \cdots, R_N, R_I, R_O, R_L, 1)$ TT factorization expresses $\tensor{W}$ as:
\begin{align*}
\tensor{W}(i_1, \cdots, i_d, i_c, i_o, i_l) =
\tensor{G}_1(i_1)
\cdot
\times
\tensor{G}_N(i_d)
\tensor{G}_I(i_c)
\times 
\cdots
\tensor{G}_O(i_o)
\times 
\cdots
\tensor{G}_L(i_l). 
\end{align*}
Where each of the factors of the decompositions $\tensor{G}_k$ are third order tensors of size $R_k \times I_k \times R_{k+1}$. 

In the experimental section~\ref{ssec:results}, we show results of \TNO trained with a Tucker, TT and CP factorization.

\paragraph{Separable Fourier Convolution} The proposed tensorization approach introduces a factorization of the weights in the spectral domain. When a CP~\cite{kolda2009tensor} is used, this induces separability over the learned kernel. We propose to make this separability explicit by not performing any channel mixing in the spectral domain and relying on the MLP introduced above to do so. The separable Spectral convolution can be thought of as a depthwise convolution performed in the Fourier domain, e.g. without any channel mixing. The mixing between channels is instead done in the spatial domain. This results in a significant reduction in the number of parameters while having minimal impact on performance (we found it necessary to increase the depth of the network, however, to ensure the network retained enough capacity).

\subsection{Multi-Grid Domain Decomposition}
\label{sec:multi-grid}
Having introduced our decomposition in the operator's parameter space, we now introduce our novel multi-grid approach to decompose the problem domain.

\textbf{Domain decomposition} is a method commonly used to parallelize classical solvers for time-dependent PDEs that is based on the principle that the solution for a fixed local region in space depends mostly on the input at the same local region \citep{Chan1994Domain}. In particular, since the time-step \(h>0\) of the numerical integrator is small, the solution \(u(x,t+h)\), for any point \(x \in D\) and \(t \in \Real_+\), depends most strongly on the points \(u(y,t)\) for all \(y \in B \big(x,r(h))\) where \(B \big( x,r(h) \big)\) denotes the ball centered at \(x\) with radius \(r(h)\). This phenomenon is easily seen for the case of the heat equation where, in one dimension, the solution satisfies
\begin{align} \nonumber
u(x,t+h) \propto \int_{-\infty}^\infty \text{exp} \left ( \frac{-(x-y)^2}{4h} \right ) u(y,t) \: \mathsf{d}y \\
\nonumber
\approx \int_{x - 4h}^{x+4h } \text{exp} \left ( \frac{-(x-y)^2}{4h} \right ) u(y,t) \: \mathsf{d}y
\end{align}
with the approximation holding since \(99.9937\%\) of the kernel's mass is contained within \(B(x,4h)\). While some results exist, there is no general convergence theory for this approach, however, its empirical success has made it popular for various numerical methods \citep{Albin2011Aspectral}. 

To exploit this localization , the domain \(D\) is split in \(q \in \Natural\) pairwise-disjoint regions \(D_1,\cdots,D_q\) so that \(D = \cup_{j=1}^q D_j\). Each region \(D_j\) is then embedded into a larger one \(Z_j \supset D_j\) so that points away from the center of \(D_j\) have enough information to be well approximated. A model can then be trained so that the approximation \(\G(a|_{Z_j})|_{D_j} \approx u|_{D_j}\) holds for all \(j \in [q]\). This idea is illustrated in Figure~\ref{fig:patching_only} where \(D=[0,1]^2\) and all \(D_j\), \(Z_j\) are differently sized squares. This allows the model to be ran fully in parallel hence its time and memory complexities are reduced linearly in \(q\).

\textbf{Multi-Grid.  }
Domain decomposition works well in classical solvers when the time step \(h > 0\) is small because the mapping \(u(\cdot,t) \mapsto u(\cdot,t+h)\) is close to the identity. However, the major advancement made by machine learning-based operator methods for PDEs is that a model can approximate the solution, in one shot, for very large times i.e. \(h > 1\). But, for larger \(h\), the size of \(Z_j\) relative to \(D_j\) must increase to obtain the same approximation accuracy, independently of model capacity. This causes any computational savings made by the decomposition approach to be lost. 

To mitigate this, we propose a multi-grid based domain decomposition approach where global information is added hierarchically at different resolutions. While our approach is inspired by the classical multi-grid method, it is not based on the V-cycle algorithm \citep{McCormick1985Multigrid}. For ease of presentation, we describe this concept when a domain \(D = \TT^2\) is uniformly discretized by \(2^s \times 2^s\) points, for some \(s \in \Natural\), but note that generalizations can readily be made. Given a final level \(L \in \Natural\), we first sub-divide the domain into $2^{2L}$ total regions each of size \(2^{s-L} \times 2^{s-L}\) and denote them \(D^{(0)}_1,\cdots,D^{(0)}_{2^{2L}}\). We call this the zeroth level. Then, around each \(D^{(0)}_j\), for any \(j \in [2^{2L}]\), we consider the square \(D^{(1)}_j\) of size \(2^{s-L+1} \times 2^{s-L+1}\) that is equidistant, in every direction, from each boundary of \(D^{(0)}_j\). We then subsample the points in \(D^{(1)}_j\) uniformly by a factor of \(\frac{1}{2}\) in each direction, making \(D^{(1)}_j\) have \(2^{s-L} \times 2^{s-L}\) points. We call this the first level. We continue this process by considering the squares \(D^{(2)}_j\) of size \(2^{s-L+2} \times 2^{s-L+2}\) around each \(D^{(1)}_j\) and subsample them uniformly by a factor of \(\frac{1}{4}\) in each direction to again yield squares with \(2^{s-L} \times 2^{s-L}\) points. The process is repeated until the \(L\)th level is reached wherein \(D^{(L)}_j\) is the entire domain subsampled by a factor of \(2^{-L}\) in each direction. The process is illustrated for the case \(L=2\) in Figure~\ref{fig:multigrid_patching}. Since we work with the torus, the region of the previous level is always at the center of the current level.  

\begin{figure}[t]
\centering
     % \subfigure[]{\includegraphics[width=0.32\textwidth]{./figure/bos_onto6.png}\label{fig:SGD_1_single}}
    \caption{\textbf{Tensorization: error in logscale as a function of the compression ratio.} We compare the tensor neural operator with an FNO with the same number of parameters (\emph{trimmed}). \textbf{We achieve over 100x compression ratio with better performance that the original FNO}}% with both Tucker, and over 400x compression ratio with CP.}
    \label{fig:compression}
    \includegraphics[width=.77\linewidth]{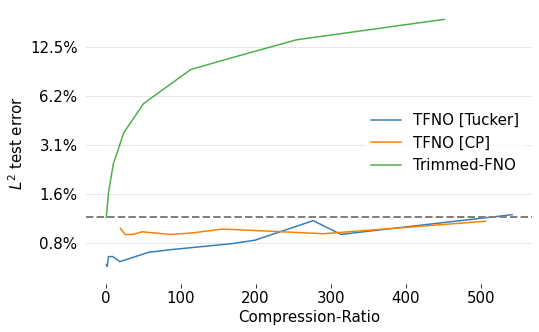}
\end{figure}

\begin{figure}
\centering
\caption{\textbf{MG-Domain Decomposition: error as a function of the domain compression ratio.} We compare \alg with different numbers of multigrid regions both with and without weight tensor compression to a full field FNO model. \textbf{We achieve over 7x input space compression, 10x parameter space compression ratios and better performance than the original FNO.}}
    \label{fig:ns_patching}
    \includegraphics[width=.77\linewidth]{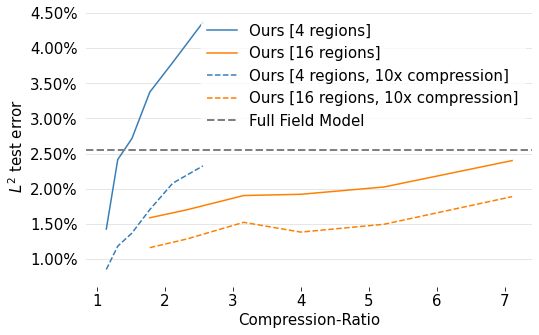}
\end{figure}

The intuition behind this method is that since the dependence of points inside a local region diminishes the further we are from that region, it is enough to have coarser information, as we go farther. We combine this multi-grid method with the standard domain decomposition approach by building appropriately padded squares \(Z^{(l)}_j\) of size \(2^{s-L} +2p \times 2^{s-L} + 2p\) around each \(D^{(l)}_j\) where \(p \in \Natural\) is the amount of padding to be added in each direction. We then take the evaluations of the input function \(a\) at each level and concatenate them as channels. In particular, we train a model so that 
\(\hat{\G}\big ( (a|_{Z_j^{(0)}},\cdots,a|_{Z_j^{(L)}} ) \big )|_{D_j^{(0)}} \approx u|_{D_j^{(0)}}.\)
Since the model only operates on each padded region separately, we reduce the total number of grid points used from $2^{2s}$ to $(2^{s-L} + 2p)^2$ and define the domain compression ratio as the quotient of these numbers.
Furthermore, note that, assuming \(a\) is \(\Real^{d_\A}\)-valued, a model that does not employ our multi-grid domain decomposition uses inputs with \(d_\A\) channels while our approach builds inputs with \((L+1)d_\A\) channels. 
In particular, the number of input  channels scales only logarithmically in the number of regions hence global information is added at very little additional cost. Indeed, FNO models are usually trained with internal widths much larger than \(d_\A\) hence the extra input channels cause almost no additional memory overhead.

%% file: tables/notation.tex
\begin{table}[]
    \centering
    \begin{tabular}{ccc}
    \toprule
    \textbf{Variable} & \textbf{Meaning} & \textbf{Dimensionality}  \\
    \midrule
    \textbf{T} & Tensor of weights in the Fourier domain & $\Cov^{\kmax \times \cdots \times \kmax \times m \times n}$\\
    $\tensor{W}$ & Weight tensor parameterizing the entire operator & $\Cov^{\kmax \times \cdots \times \kmax \times n \times n \times 2^{d-1} L}$\\
    $\A$ & Input function space & Infinite \\
    $\U$ & output function space & Infinite \\
    $a$ & Input function & Infinite \\
    $u$ & Output function & Infinite \\
    $D_\A$ & Domain of function a& $d$ \\
    $D_\U$ & Domain of function u& $d$ \\
    $d_\A$ & Dimension of the co-domain of the input functions & $1$ \\
    $d_\U$ & Dimension of the co-domain of the output functions& $1$ \\
    $\F$ & Fourier transform & Infinite \\
    $\F^{-1}$ & Fourier transform & Infinite \\
    $L$ & Number of integral operation layers & In $\Natural$ \\
    $l$ & Layer index & Between $1$ and $L$ \\
    %$\P$ & Point-wise lifting operator & Infinite \\
    %$\Q$ & Point-wise projeotion operator & Infinite \\
    %$\W$ & Point-wise residual connection operator & Infinite \\
    $\sigma$ & Point-wise activation operation& Infinite \\
    %$\K$ & Integral operator & Infinite \\
    %$\kappa$ & Kernel of Integral operator & Infinite \\
    $b$ & Bias vector &  \\
    $v$ & Function at each layer & Infinite \\
    $\kmax$ & Number of kept frequencies in Fourier space & Between $1$ and $\frac{1}{2} \min \{s_1,\cdots,s_d\}$
    \\
    \bottomrule
    \end{tabular}
    \caption{\textbf{Table of notation}}
    \label{tab:notation}
\end{table}

%% file: sections/experiments-setting.tex
\section{Experiments}

In this section, we first introduce the data, experimental setting and implementation details before empirically validating our approach through thorough experiments and ablations.

\subsection{Data.}\label{sec:app_data}
We experiment on a dataset of 10K training samples and 2K test samples of the two-dimensional Navier-Stokes equation with Reynolds number 500.
We also experiment with the one-dimensional viscous Burgers' equation.

\paragraph{Navier-Stokes.}
We consider the vorticity form of the two-dimensional Navier-Stokes equation,
\begin{align}
    \label{eq:ns}
    \begin{split}
        \partial_t \omega + \nabla^\perp &\phi \cdot \omega = \frac{1}{\text{Re}} \Delta \omega + f, \quad \:\: x \in\mathbb{T}^2, \: t \in (0,T]  \\
        - \Delta \phi &= \omega, \quad \int_{\mathbb{T}^2} \phi = 0, \qquad x \in \mathbb{T}^2, \: t \in (0,T]
    \end{split}
\end{align}
with initial condition \(\omega(0,\cdot) = 0\) where \(\mathbb{T}^2 \cong [0,2\pi)^2\) is the torus, \(f \in \dot{L}^2(\mathbb{T}^2;\Real)\) is a forcing function, and \(\text{Re} > 0\) is the Reynolds number. Then \(\omega(t,\cdot) \in \dot{H}^s(\mathbb{T}^2;\Real)\) for any \(t \in (0,T]\) and \(s > 0\), is the unique weak solution to \eqref{eq:ns} \citep{temam1988infinite}. We consider the non-linear operator mapping \(f \mapsto \omega(T,\cdot)\) with \(T=5\) and fix the Reynolds number \(\text{Re} = 500\). We define the Gaussian measure \(\mu = \mathcal{N}(0,C)\) on the forcing functions where we take the covariance \(C = 27 (-\Delta + 9I)^{-4}\), following the setting in \citep{de2022cost}. Input data is obtained by generating i.i.d. samples from \(\mu\) by a KL-expansion onto the eigenfunctions of \(C\) \citep{powell2014AnIntorduction}. Solutions to \eqref{eq:ns} are then obtained by a pseudo-spectral scheme \citep{chandler2013invariant}.

\paragraph{Burgers' Equation.}
\label{sec:app_burgers}
We consider the one-dimensional Burgers' equation on the torus,
\begin{align}
    \label{eq:burgers}
    \begin{split}
    \partial_t u + u u_x &= \nu u_{xx}, \qquad x \in \TT, \: t \in (0,T] \\
    u|_{t=0} &= u_0, \qquad \quad  x \in \TT
    \end{split}
\end{align}
for initial condition \(u_0 \in L^2(\TT;\Real)\) and viscosity \(\nu > 0\). Then \(u(t,\cdot) \in H^s (\TT;\Real)\), for any \(t \in \Real_+\) and \(s > 0\), is the unique weak solution to \ref{eq:burgers} \citep{evans2010partial}. We consider the non-linear operator \(u_0 \mapsto u(T,\cdot) \) with \(T = 0.5 \text{ or } 1\) and fix \(\nu = 0.01\). We define the Gaussian measure \(\mu = \mathcal{N}(0,C)\) where we take the covariance \(C = 3^{5/2} (- \frac{d^2}{dx^2} + 9I)^{-3}\). Input data is obtained by generating i.i.d. samples from \(\mu\) by a KL-expansion onto the eigenfunctions of \(C\). Solutions to \eqref{eq:burgers} are then obtained by a pseudo-spectral solver using Heun's method. We use 8K samples for training and 2K for testing.

\subsection{Implementation details}
\paragraph{Implementation} We use PyTorch~\cite{pytorch} for implementing all the models. The tensor operations are implemented using TensorLy~\cite{tensorly} and TensorLy-Torch~\cite{tensorly-torch}. Our code was released under the permissive MIT license, as a Python package that is well-tested and comes with extensive documentation, to encourage and facilitate downstream scientific applications. It is available at \url{https://github.com/neuraloperator/neuraloperator}.

\paragraph{Hyper-parameters} We train all models via gradient backpropagation using a mini-batch size of $16$, the Adam optimizer, with a learning rate of $1e^{-3}$, weight decay of $1e^{-4}$, for $500$ epochs, decreasing the learning rate every $100$ epochs  by a factors of $\frac{1}{2}$. The model width is set in all cases to $64$ except when specified otherwise (for the Trimmed \FNO), meaning that the input was first lifted (with a linear layer) from the number of input channels to that width. The projection layer projects from the width to $256$ and a prediction linear layer outputs the predictions. $10000$ samples were used for training, as well as a separate set of $2000$ samples for testing. All experiments are done on a NVIDIA Tesla V100 GPU.

To disentangle the effect of each of our components, the comparisons between the original FNO, the MG-FNO, \TOP, and the \alg were conducted in the same setting, with a mini-batch size of 32, modes of 42 and 21 for the height and width, respectively, and an operator width of 64. 

For the comparison between our best models, we use all the modes (64 and 32) and a mini-batch size of 16, which leads to improved performance for all models but longer training times. For each comparison, the same setting and hyper-parameters were used for all models.

\paragraph{Training the operator.}
Since \alg predicts local regions which are then stitched together to form a global function without any communication, aliasing effects can occur where one output prediction does not flow smoothly into the next. To prevent this, we train our model using the \(H^1\) Sobolev norm \citep{czarnecki2017sobolev,li2021markov}. By matching derivatives, training with this loss prevents any discontinuities from occurring and the output prediction is smooth.

%% file: sections/experiments-results.tex
\subsection{Experimental results}\label{ssec:results}

In this section, we compare our approach with both the regular FNO~\cite{FNO} and the Factorized-FNO~\cite{tran2023factorized}, which separately applied FFT along each mode before combining the results. In all cases, our approach achieves superior performance with a fraction of the parameters, as can be seen in Table~\ref{tab:comparison}.

\begin{figure*}[tb]
    \centering
    \includegraphics[width=0.49\linewidth]{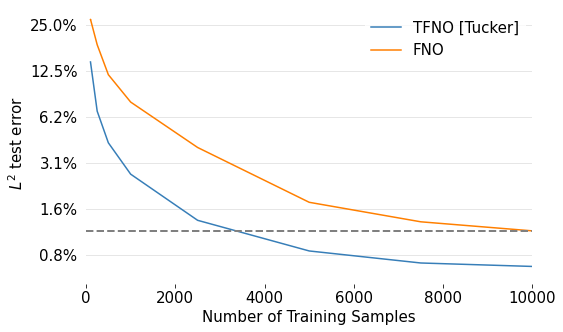}
    \hfill
    \includegraphics[width=0.49\linewidth]{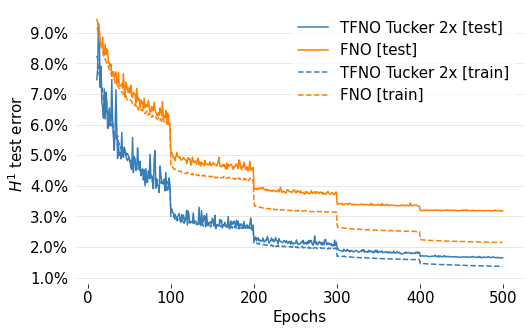}
    \caption{\textbf{Error as a function of the number of training samples (left) and training VS testing loss.} We compare \TOP with a regular \FNO. Note that on the left we show the testing $L^2$ error while, for training, the $H^1$ loss is used and that is compared with the $H^1$ test error on the right. Our approach generalizes better while requiring fewer training samples.}
    \label{fig:low-data-regime}
\end{figure*}

\paragraph{Tensorizing: better compression.}
In Figure~\ref{fig:compression}, we show the performance of our approach (TNO) compared to the original \FNO, for varying compression ratios. In the Trimmed-\FNO, we adjust the width in order to match the number of parameters in our TNO. We focus on the width of the network as it was shown to be the most important parameter~\citep{de2022cost}. Our method massively outperforms the Trimmed-\FNO at every single fixed parameter amount. Furthermore, even for very large compression ratios, our FNO outperforms the full-parameter FNO model. This is likely due to the regularizing effect of the tensor factorization on the weight, showing that many of the ones in the original model are redundant. 

\paragraph{Tensorizing: better generalization.}
 Figure~\ref{fig:low-data-regime} (left) shows that our TNO generalizes better with less training samples. Indeed, at every fixed amount of training samples, the TNO massively outperforms the full-parameter FNO model. Even when only using half the samples, our TNO outperforms the FNO trained on the full dataset. Furthermore, Figure~\ref{fig:low-data-regime} (right) shows that our TNO overfits significantly less than FNO, demonstrating the regularizing effect of the tensor decomposition.  This result is invaluable in the \PDE setting where very few training samples are typically available due to the high computational cost of traditional \PDE solvers.
    
 \paragraph{Multi-Grid Domain Decomposition.}
In Table~\ref{tab:mgtfno}, we compare our \alg with the baseline FNO and the \TOP, respectively. \alg enables compressing both the weight tensor but also the input domain. On the other hand, preserving resolution invariance requires padding the patches, which decreases performance, resulting in a tradeoff between input domain compression and prediction accuracy. 

We also show the impact of multi-grid domain decomposition on performance in Figure~\ref{fig:ns_patching}. We find that lower compression ratios (corresponding to a larger amount of padding in the decomposed regions) perform better which is unsurprising since more information is incorporated into the model. More surprisingly, we find that using a larger number of regions (16) performs consistently better than using a smaller number (4) and both can outperform the full-field FNO. This can be due to the fact that: i) the domain decomposition acts as a form of data augmentation, exploiting the translational invariance of the \PDE and more regions yield larger amounts of data, and ii)  the output space of the model is simplified since a function can have high frequencies globally but may only have low frequencies locally. Consistently, we find that the tensor compression in the weights acts as a regularizer and improves performance across the board.

\begin{table}[]
    \label{tab:comparison}
    \caption{\textbf{Comparing the performance of \alg with previous works on Navier-Stokes}. Our method achieves superior performance with a fraction of the parameters while largely compressing the weights (\TOP) and the input-domain (\alg).}
    \centering
    \begin{tabular}{lllll}
    \toprule
    \textbf{Method} & \textbf{$L^2$ test error (\%)} & \textbf{\# Params} & \textbf{Model CR} & \textbf{Input CR} 
    \normalsize
    \\
    \midrule
    \textbf{\FNO}~\cite{FNO} & 1.34\% & 67M & - & -
    \\
    \textit{FFNO}~\cite{tran2023factorized} &	1.15 \% & 1M	 & $67 \times$ & - 
    \\
    \midrule
    \textbf{\TOP} (CP) & 0.29\% &  890K &  $75 \times$ & - 
    \\
    \textbf{\TOP} (CP) & 0.47\%  & 447K &  $150 \times$ & - 
    \\
    \midrule
    \textbf{\alg} (CP) & 0.49 \% & 447K &  $40 \times$ & $1.9 \times$ 
    \\
    \textbf{\alg} (Tucker) & 0.42 \% & 447K &  $19 \times$ & $1.9 \times$
    \\
    \bottomrule
    \end{tabular}
\end{table}

\paragraph{Architectural improvements to the backbone} In addition to the ablation performed on our \alg, we also investigate architectural improvements to the \FNO backbone, see Sec~\ref{sec:architecture} for details. In particular, we find that, while instance normalization decreases performance, layer normalization helps, especially when used in conjunction with a pre-activation. Adding an MLP similarly improves performance, we found that a bottleneck (expansion factor of 0.5) works well in practice, resulting in an absolute improvement of $0.87\%$ in relative $L^2$ error. We found the ordering of normalization, activation, and weights (including preactivation), did not have a significant impact on performance. Finally, when not using multi-grid domain decomposition, the inputs are periodic and padding is not necessary. In that case, not padding the input improves performance. We use all these improvements for the backbone of the best version of our \alg, Fig~\ref{fig:comparison_best} where we show that our improved backbone significantly outperforms the original FNO, while our approach significantly outperforms both, with a small fraction of the parameters, opening the door to the application of \alg to high-resolution problems. 

% TODO: complete
\begin{table}[]
    \label{tab:depth}
    \caption{\textbf{Impact of our architectural improvements}. }
    \centering
    \begin{tabular}{llllrr}
    \toprule
    \textbf{Method} & \textbf{Layers} & \textbf{$L^2$ test error} & \textbf{$H^1$ test error} &  \textbf{\# Params} & \textbf{Model CR} 
    \normalsize
    \\
    \midrule
    \textbf{\FNO}~\cite{FNO}  &  4 & $1.34 \%$ & $3.78 \%$ & 67,142,657 & -
    \\
    \textbf{\FNO}~\cite{FNO}  &  6 & $0.90 \%$ & $2.59 \%$ & 100,705,409 & $0.7 \times$
    \\
    \textbf{\FNO}~\cite{FNO}  &  8 & $0.73 \%$ & $2.09 \%$ & 134,268,161 & $0.5\times$
    \\
    \midrule
    \textbf{\TOP} (CP) & 4 & $0.47 \%$ & $1.20 \%$ & 447,105 &  $150 \times$
    \\
    \textbf{\TOP} (CP) & 6 & $0.27 \%$ & $0.74 \%$ & 662,081 &  $101\times$
    \\
    \textbf{\TOP} (CP) & 8 & $0.22 \%$ & $0.59 \%$ & 877,057 & $77\times$
    \\
    \bottomrule
    \end{tabular}
\end{table}

\subsection{Ablation studies}\label{sec:app_ablation}
In this section, we further study the properties of our model through ablation studies. We first look at how \TOP suffers less from overfitting thanks to the low-rank constraints before comparing its performance with various tensor decompositions. Finally, we perform ablation studies for our multi-grid domain decomposition on Burger's equation.

\subsubsection{Resolution invariance}
\TOP is resolution invariant, meaning that it can be trained on one resolution and tested on a different one. To illustrate this, we show zero-shot super-resolution results: we trained our best model (Table~\ref{tab:best}) on images of resolution $128 \times 128$ and tested it on unseen samples at higher resolutions ($256 \times 256$ and $512 \times 512$), Table~\ref{tab:resolution-invariance}. As can be seen, our method does as well on unseen, higher-resolution unseen testing samples as it does on the training resolution, confirming the resolution invariance property of our neural operator. 

\begin{table}[]
    \centering
    \caption{\textbf{Resolution invariance of \TOP.} Since the model is an operator, it is resolution invariant. In particular, here, we trained our model in resolution $128 \times 128$ and test it on unseen samples in various resolutions and show it generalizes, with virtually no loss of performance to higher resolutions unseen during training.}
    \label{tab:resolution-invariance}
    \begin{tabular}{lllllllll}
    \toprule
    \multirow{2}{*}{\textbf{Method}}
                    & \multicolumn{2}{c}{\textbf{$128 \times 128$}} 
                    & \multicolumn{2}{c}{\textbf{$256 \times 256$}}
                    & \multicolumn{2}{c}{\textbf{$512 \times 512$}}
                    & \multicolumn{2}{c}{\textbf{$1024 \times 1024$}}
    \\
    \cmidrule(lr){2-3} \cmidrule(lr){4-5} \cmidrule(lr){6-7} \cmidrule(lr){8-9}
    & \small \textbf{$L^2$ error} & \small \textbf{$H^1$ error} 
    & \small \textbf{$L^2$ error} & \small \textbf{$H^1$ error} 
    & \small \textbf{$L^2$ error} & \small \textbf{$H^1$ error} 
    & \small \textbf{$L^2$ error} & \small \textbf{$H^1$ error} 
    \normalsize
    \\
    \midrule
    \textbf{CP \TOP} 
    & $0.3\%$ & $0.87 \%$% 128
    & $0.3\%$ & $0.93 \%$ % 256
    & $0.3\%$ & $0.93 \%$% 512
    & $0.3\%$ & $0.93 \%$ % 1024
    \\
    \textbf{CP \alg} 
    & $0.49\%$ & $1.2 \%$% 128
    & $0.49\%$ & $1.3 \%$ % 256
    & $0.49\%$ & $1.5 \%$% 512
    & $0.49\%$ & $1.6 \%$ % 1024
    \\
    \bottomrule
\end{tabular}
\end{table}

\subsubsection{Training on higher-resolution with Multi-grid}
One important advantage of our multi-grid domain decomposition is that it enables training much larger models on large inputs by distributing over patches. We demonstrate this, by training on larger resolution (512x512 discretization) and using the largest FNO and TFNO that fits in memory, on a V100 GPU. For the original FNO, this corresponds to a width of 12, first row in table~\ref{tab:training-512}. We then compare its performance with the multigrid approach with a neural operator as large as fits into the same V100 GPUs i.e. each width in the table has been optimized to be as large as memory allows. As we can see, our approach allows to fit a larger model and reaches a much lower relative $L^2$ error.

\begin{table}[]
    \centering
    \caption{\textbf{Training on 512x512}. Multi-grid domain decomposition allows us to fit larger models into memory by distributing patches in the domain space, thus reaching a lower relative error.}
    \label{tab:training-512}
    % \resizebox{1\linewidth}{!}{%
    \begin{tabular}{l l l l l} 
    \toprule
    Model & Width & Patches & Padding & $L^2$ error  \\ 
    \midrule
    FNO & 12 & 0 & 0 & 6.1 \\% \\\\ 
    \midrule
    MG-FNO & 42 & 4 & 70 & 2.9 \\% \\\\ 
    \midrule
    MG-FNO &66 & 4 & 53 & 2.4 \\% \\\\ 
    \midrule
    MG-FNO & 88 & 16 & 40 & 1.8 \\% \\\\ \\hline
    Tucker MG-TFNO &80& 16 & 46 & 1.3 \\% \\\\ 
    \bottomrule
    \end{tabular}
    % }
\end{table}

\subsubsection{Overfitting and Low-Rank Constraint}
Here, we show that lower ranks (higher compressions) lead to reduced overfitting. In Figure~\ref{fig:overfitting-study-tucker}, we show the training and testing $H^1$ errors for our TOP with Tucker decomposition at varying compression ratios (2x, 49x and 172x). We can see how, while the test error does not vary much, the gap between training and test errors reduces as we decrease the rank. As we can see, while being the most flexible, Tucker does not perform as well at higher compression ratios. In those extreme cases, CP and Tensor-Train lead to lower errors.

\begin{figure}[htb]
    \subfigure[\textbf{Train VS Test error over time for a TOP with a CP factorization).}]{%
    \label{fig:overfitting-cp}
    \includegraphics[width=0.48\linewidth]{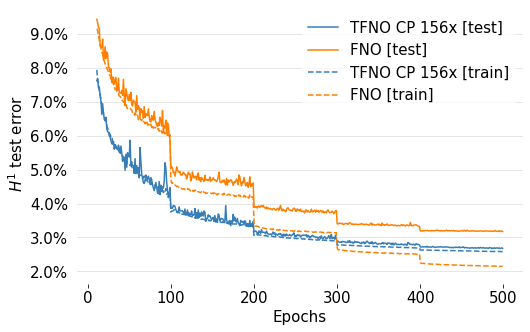}
    }
    \hfill
    \subfigure[\textbf{Train VS Test error over time for a TOP with a TT factorization.}]{%
    \label{fig:overfitting-tt}
    \includegraphics[width=0.48\linewidth]{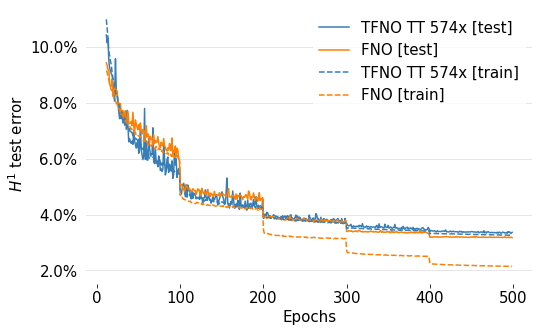}
    }
    \caption{\textbf{Train/test curve for a TOP-CP (\ref{fig:overfitting-cp}) and  TOP-TT (\ref{fig:overfitting-tt})}}
\end{figure}

\subsubsection{Tensor-Train and TOP}
Our approach is independent of the choice of tensor decomposition. We already showed how Tucker is most flexible and works well across all ranks. We also showed that while memory demanding for high rank, a CP decomposition leads to better performance and low rank. Our method can also be used in conjunction with other decompositions, such as tensor-train. To illustrate this, we show the convergence behavior of TNO with a Tensor-Train decomposition for a compression ratio of 178, figure~\ref{fig:overfitting-tt}. 

\input{tables/comparison_decompositions.tex}
We also compare in Table~\ref{tab:comparison} our \TOP with different tensor decompositions.

\subsubsection{Decomposing domain and weights: \alg.}
Tensorization and  multi-grid domain decomposition not only improve performance individually, but their advantages compound and lead to a strictly better algorithm that scales well to higher-resolution data by decreasing the number of parameters in the model as well as the size of the inputs thereby improving performance as well as memory and computational footprint. Table~\ref{tab:overall_results} compares FNO with Tensorization alone, multi-grid domain decomposition alone, and our joint approach combining the two, \alg. In all cases, for $\kmax$, we keep $40$ Fourier coefficients for height and $24$ for the width and use an operator width of $64$. Our results imply that, under full parallelization, the memory footprint of the model's inference can be reduced by $7\times$ and the size of its weights by $10\times$ while also improving performance.

\input{tables/overall_results.tex}

Consistently with our other experiments, we find that the tensor compression in the weights acts as a regularizer and improves performance across the board. Our results imply that, under full parallelization, the memory footprint of the model's inference can be reduced by $7\times$ and its weight size by $10\times$ while also improving performance.

\subsubsection{Burgers' Equation}

\begin{figure}[htb]
    \centering
    \includegraphics[width=0.49\linewidth]{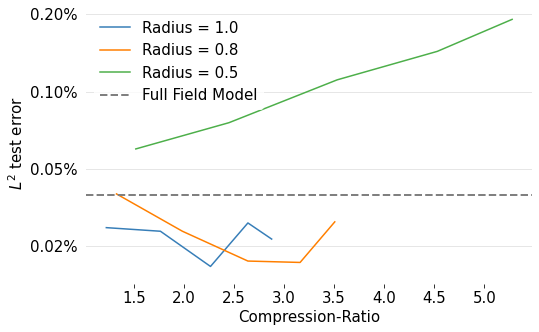}
    \hfill
    \includegraphics[width=0.49\linewidth]{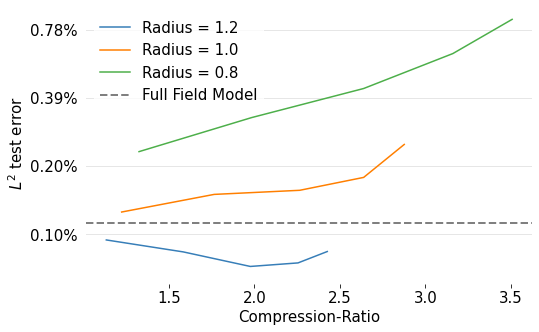}
    \caption{\textbf{Error on Burgers' equation with \(T=0.5\) (left) and \(T=1\) (right) as a function of domain compression ratio using standard domain decomposition without our multi-grid approach.} We evaluate the performance of the standard domain decomposition approach. The radius indicates the size, in physical space, of the padding added to each region.}
    \label{fig:burgers}
\end{figure}

We test the efficacy of the standard domain decomposition approach by training on two separate Burgers problems: one with a final time \(T=0.5\) and one with \(T=1\). As described in Section~\ref{sec:multi-grid}, we expect that for \(T=1\), each region requires more global information thus significantly more padding need to be used in order to reach the same error. 
The results of Figure~\ref{fig:burgers} indeed confirm this. The domain compression ratios needed for the approach to reach the performance of the full-field model are higher, indicating the need for incorporating global information. These results motivate our multi-grid domain decomposition approach.

%% file: tables/comparison_decompositions.tex
\begin{table}[]
    \label{tab:comparison}
    \caption{\textbf{Relative $L^2$ test error of our \alg approach for different tensor decompositions}. We empirically found that Tucker works best for small compression ratio, CP excels at large compression ratio ($\approx 100 \times$) but becomes computationally heavy for smaller ones. TT tends to be unstable at low-compression ratios but preserves a good performance for extreme compression ratio ($> 500 \times$).}
    \centering
    % \begin{tabular*}{\textwidth}{@{\extracolsep{\fill}}lllll}
    \begin{tabular}{llll}
    \toprule
    \textbf{Method} & \small \textbf{$L^2$ test error} & \small \textbf{\# Params} & \small \textbf{Model CR}
    \normalsize
    \\
    \midrule
    \textbf{\FNO}~\cite{FNO}  & $1.12\%$ & $67$~M  & $0\times$ % 67142657
    \\
    \midrule
    \textbf{\TOP} [Tucker] & $0.37\%$ & $28$~M & $2.3\times$ %28518273
    \\
    \textbf{\TOP} [CP] & $0.46\%$ & $808$~K & $83\times$ %808577
    \\
    \textbf{\TOP} [TT] & $1.18\%$ & $117$~K & $574\times$ % 116993
    \\
    \bottomrule
    \end{tabular}
\end{table}

%% file: tables/overall_results.tex
\begin{table}[]
    \centering
    \caption{\textbf{Ablation comparing the performance on the relative $L^2$ test error of our \alg approach, compared with its parts \TOP and MG-FNO and the regular FNO, on Navier-Stokes.} CR stands for compression ratio. 
    Tensorization and multi-grid domain decomposition both individually improve performance while enabling space savings. The two techniques combined lead to further improvements, enabling large compression for both input and parameter, while outperforming regular \FNO.
    }
    \label{tab:overall_results}
    % \resizebox{1\linewidth}{!}{%
    % \begin{tabular*}{\textwidth}{@{\extracolsep{\fill}}lllll}
    \begin{tabular}{lllll}
    \toprule
    \textbf{Method} & \small \textbf{$L^2$ test error} & \small \textbf{\# Params} & \small \textbf{Model CR} & \small \textbf{Domain CR}
    \normalsize
    \\
    \midrule
    \textbf{FNO}~\citep{FNO}  & $2.54\%$ & $58$~M  & $0\times$ &  $0\times$
    \\
    \midrule
    \textbf{TFNO} [Tucker] & $1.39\%$ & $41$~M & $1.5\times$ &  $0\times$
    \\
    \textbf{TFNO} [CP] & $2.24\%$ & $130$~K & $\bf{482\times}$ &  $0\times$
    % \textbf{TOP} [Tucker] & $2.97\%$ & $139$~K & $\bf{451\times}$ &  $0\times$
    \\
    \midrule 
    \textbf{MG-FNO} & $1.43\%$ & $58$~M & $ 0 \times$ &  $1.4 \times$ 
    \\
    \midrule
    \textbf{MG-TFNO} [Tucker] & $\bf{0.85\%}$ & $5.5$~M & $10 \times$ &  $1.78 \times$ 
    \\
    \textbf{MG-TFNO} [Tucker] & $1.89\%$ & $5.5$~M & $10 \times$ &  $\bf{7 \times}$ 
    \\
    \bottomrule
    \end{tabular}
    % }
\end{table}

%% file: sections/conclusions.tex
\section{Conclusion}
In this work, we introduced i) a novel tensor operator (\TOP) as well as a multi-grid domain decomposition approach which together form \alg, ii) an operator model that outperforms the \FNO with a fraction of the parameters and memory complexity requirements, and iii) architectural improvements to the \FNO. Our method scales better, generalizes better, and requires fewer training samples to reach the same performance; while the multi-grid domain decomposition enables parallelism over huge inputs. This paves the way to applications on very high-resolution data and in our future work, we plan to deploy \alg to large-scale weather forecasts for which existing deep learning models are prohibitive.

%% file: main.bbl
\begin{thebibliography}{61}
\providecommand{\natexlab}[1]{#1}
\providecommand{\url}[1]{\texttt{#1}}
\expandafter\ifx\csname urlstyle\endcsname\relax
  \providecommand{\doi}[1]{doi: #1}\else
  \providecommand{\doi}{doi: \begingroup \urlstyle{rm}\Url}\fi

\bibitem[Adler \& Oktem(2017)Adler and Oktem]{Adler2017}
Jonas Adler and Ozan Oktem.
\newblock Solving ill-posed inverse problems using iterative deep neural
  networks.
\newblock \emph{Inverse Problems}, nov 2017.

\bibitem[Albin \& Bruno(2011)Albin and Bruno]{Albin2011Aspectral}
Nathan Albin and Oscar~P. Bruno.
\newblock A spectral fc solver for the compressible navier–stokes equations
  in general domains i: Explicit time-stepping.
\newblock \emph{Journal of Computational Physics}, 230\penalty0 (16):\penalty0
  6248--6270, 2011.

\bibitem[Ba et~al.(2016)Ba, Kiros, and Hinton]{layernorm}
Jimmy~Lei Ba, Jamie~Ryan Kiros, and Geoffrey~E Hinton.
\newblock Layer normalization.
\newblock \emph{arXiv preprint arXiv:1607.06450}, 2016.

\bibitem[Bhatnagar et~al.(2019)Bhatnagar, Afshar, Pan, Duraisamy, and
  Kaushik]{bhatnagar2019prediction}
Saakaar Bhatnagar, Yaser Afshar, Shaowu Pan, Karthik Duraisamy, and Shailendra
  Kaushik.
\newblock Prediction of aerodynamic flow fields using convolutional neural
  networks.
\newblock \emph{Computational Mechanics}, pp.\  1--21, 2019.

\bibitem[Bhattacharya et~al.(2020)Bhattacharya, Hosseini, Kovachki, and
  Stuart]{bhattacharya2020model}
Kaushik Bhattacharya, Bamdad Hosseini, Nikola~B Kovachki, and Andrew~M Stuart.
\newblock Model reduction and neural networks for parametric pdes.
\newblock \emph{arXiv preprint arXiv:2005.03180}, 2020.

\bibitem[Blanusa et~al.(2022)Blanusa, L{\'o}pez-Zurita, and
  Rasp]{blanusa2022role}
Mackenzie~L Blanusa, Carla~J L{\'o}pez-Zurita, and Stephan Rasp.
\newblock The role of internal variability in global climate projections of
  extreme events.
\newblock \emph{arXiv preprint arXiv:2208.08275}, 2022.

\bibitem[Bulat et~al.(2020{\natexlab{a}})Bulat, Kossaifi, Tzimiropoulos, and
  Pantic]{bulat2020incremental}
Adrian Bulat, Jean Kossaifi, Georgios Tzimiropoulos, and Maja Pantic.
\newblock Incremental multi-domain learning with network latent tensor
  factorization.
\newblock In \emph{Proceedings of the AAAI Conference on Artificial
  Intelligence}, volume~34, pp.\  10470--10477, 2020{\natexlab{a}}.

\bibitem[Bulat et~al.(2020{\natexlab{b}})Bulat, Kossaifi, Tzimiropoulos, and
  Pantic]{bulat2020toward}
Adrian Bulat, Jean Kossaifi, Georgios Tzimiropoulos, and Maja Pantic.
\newblock Toward fast and accurate human pose estimation via soft-gated skip
  connections.
\newblock In \emph{2020 15th IEEE International Conference on Automatic Face \&
  Gesture Recognition}, 2020{\natexlab{b}}.

\bibitem[Chan \& Mathew(1994)Chan and Mathew]{Chan1994Domain}
Tony~F. Chan and Tarek~P. Mathew.
\newblock Domain decomposition algorithms.
\newblock \emph{Acta Numerica}, 3:\penalty0 61–143, 1994.

\bibitem[Chandler \& Kerswell(2013)Chandler and
  Kerswell]{chandler2013invariant}
Gary~J. Chandler and Rich~R. Kerswell.
\newblock Invariant recurrent solutions embedded in a turbulent two-dimensional
  kolmogorov flow.
\newblock \emph{Journal of Fluid Mechanics}, 722:\penalty0 554–595, 2013.

\bibitem[Cordonnier et~al.(2020)Cordonnier, Loukas, and
  Jaggi]{cordonnier2020multi}
Jean-Baptiste Cordonnier, Andreas Loukas, and Martin Jaggi.
\newblock Multi-head attention: Collaborate instead of concatenate.
\newblock \emph{arXiv preprint arXiv:2006.16362}, 2020.

\bibitem[Courant et~al.(1928)Courant, Friedrichs, and Lewy]{courant1928uber}
R.~Courant, K.~Friedrichs, and H.~Lewy.
\newblock Uber die partiellen differenzengleichungen der mathematischen physik.
\newblock \emph{Mathematische annalen}, 100\penalty0 (1):\penalty0 32--74,
  1928.

\bibitem[Czarnecki et~al.(2017)Czarnecki, Osindero, Jaderberg, Swirszcz, and
  Pascanu]{czarnecki2017sobolev}
Wojciech~M Czarnecki, Simon Osindero, Max Jaderberg, Grzegorz Swirszcz, and
  Razvan Pascanu.
\newblock Sobolev training for neural networks.
\newblock \emph{Advances in Neural Information Processing Systems}, 30, 2017.

\bibitem[Dao et~al.(2022)Dao, Chen, Sohoni, Desai, Poli, Grogan, Liu, Rao,
  Rudra, and R{\'e}]{dao2022monarch}
Tri Dao, Beidi Chen, Nimit~S Sohoni, Arjun Desai, Michael Poli, Jessica Grogan,
  Alexander Liu, Aniruddh Rao, Atri Rudra, and Christopher R{\'e}.
\newblock Monarch: Expressive structured matrices for efficient and accurate
  training.
\newblock In \emph{International Conference on Machine Learning}, pp.\
  4690--4721. PMLR, 2022.

\bibitem[De~Hoop et~al.(2022)De~Hoop, Huang, Qian, and Stuart]{de2022cost}
Maarten De~Hoop, Daniel~Zhengyu Huang, Elizabeth Qian, and Andrew~M Stuart.
\newblock The cost-accuracy trade-off in operator learning with neural
  networks.
\newblock \emph{arXiv preprint arXiv:2203.13181}, 2022.

\bibitem[Dosovitskiy et~al.(2020)Dosovitskiy, Beyer, Kolesnikov, Weissenborn,
  Zhai, Unterthiner, Dehghani, Minderer, Heigold, Gelly,
  et~al.]{dosovitskiy2020image}
Alexey Dosovitskiy, Lucas Beyer, Alexander Kolesnikov, Dirk Weissenborn,
  Xiaohua Zhai, Thomas Unterthiner, Mostafa Dehghani, Matthias Minderer, Georg
  Heigold, Sylvain Gelly, et~al.
\newblock An image is worth 16x16 words: Transformers for image recognition at
  scale.
\newblock \emph{arXiv preprint arXiv:2010.11929}, 2020.

\bibitem[Esmaeilzadeh et~al.(2020)Esmaeilzadeh, Azizzadenesheli, Kashinath,
  Mustafa, Tchelepi, Marcus, Prabhat, Anandkumar,
  et~al.]{esmaeilzadeh2020meshfreeflownet}
Soheil Esmaeilzadeh, Kamyar Azizzadenesheli, Karthik Kashinath, Mustafa
  Mustafa, Hamdi~A Tchelepi, Philip Marcus, Mr~Prabhat, Anima Anandkumar,
  et~al.
\newblock Meshfreeflownet: A physics-constrained deep continuous space-time
  super-resolution framework.
\newblock In \emph{SC20: International Conference for High Performance
  Computing, Networking, Storage and Analysis}, pp.\  1--15. IEEE, 2020.

\bibitem[Evans(2010)]{evans2010partial}
Lawrence~C. Evans.
\newblock \emph{Partial differential equations}.
\newblock American Mathematical Society, 2010.

\bibitem[Guibas et~al.(2021)Guibas, Mardani, Li, Tao, Anandkumar, and
  Catanzaro]{guibas2021adaptive}
John Guibas, Morteza Mardani, Zongyi Li, Andrew Tao, Anima Anandkumar, and
  Bryan Catanzaro.
\newblock Adaptive fourier neural operators: Efficient token mixers for
  transformers.
\newblock \emph{arXiv preprint arXiv:2111.13587}, 2021.

\bibitem[Guo et~al.(2016)Guo, Li, and Iorio]{guo2016convolutional}
Xiaoxiao Guo, Wei Li, and Francesco Iorio.
\newblock Convolutional neural networks for steady flow approximation.
\newblock In \emph{Proceedings of the 22nd ACM SIGKDD International Conference
  on Knowledge Discovery and Data Mining}, 2016.

\bibitem[Gupta et~al.(2021)Gupta, Xiao, and Bogdan]{gupta2021multiwavelet}
Gaurav Gupta, Xiongye Xiao, and Paul Bogdan.
\newblock Multiwavelet-based operator learning for differential equations.
\newblock \emph{Advances in Neural Information Processing Systems},
  34:\penalty0 24048--24062, 2021.

\bibitem[Gusak et~al.(2019)Gusak, Kholiavchenko, Ponomarev, Markeeva,
  Blagoveschensky, Cichocki, and Oseledets]{Gusak_2019_ICCV}
Julia Gusak, Maksym Kholiavchenko, Evgeny Ponomarev, Larisa Markeeva, Philip
  Blagoveschensky, Andrzej Cichocki, and Ivan Oseledets.
\newblock Automated multi-stage compression of neural networks.
\newblock Oct 2019.

\bibitem[He et~al.(2016)He, Zhang, Ren, and Sun]{he2016identity}
Kaiming He, Xiangyu Zhang, Shaoqing Ren, and Jian Sun.
\newblock Identity mappings in deep residual networks.
\newblock In \emph{Computer Vision--ECCV 2016: 14th European Conference,
  Amsterdam, The Netherlands, October 11--14, 2016, Proceedings, Part IV 14},
  pp.\  630--645. Springer, 2016.

\bibitem[Ioffe \& Szegedy(2015)Ioffe and Szegedy]{batchnorm}
Sergey Ioffe and Christian Szegedy.
\newblock Batch normalization: Accelerating deep network training by reducing
  internal covariate shift.
\newblock In \emph{International conference on machine learning}, pp.\
  448--456. pmlr, 2015.

\bibitem[Janzamin et~al.(2019)Janzamin, Ge, Kossaifi, Anandkumar,
  et~al.]{janzamin2019spectral}
Majid Janzamin, Rong Ge, Jean Kossaifi, Anima Anandkumar, et~al.
\newblock Spectral learning on matrices and tensors.
\newblock \emph{Found. and Trends{\textregistered} in Mach. Learn.},
  12\penalty0 (5-6):\penalty0 393--536, 2019.

\bibitem[Kim et~al.(2016)Kim, Park, Yoo, Choi, Yang, and
  Shin]{yong2016compression}
Yong{-}Deok Kim, Eunhyeok Park, Sungjoo Yoo, Taelim Choi, Lu~Yang, and Dongjun
  Shin.
\newblock Compression of deep convolutional neural networks for fast and low
  power mobile applications.
\newblock 2016.

\bibitem[Kolda \& Bader(2009)Kolda and Bader]{kolda2009tensor}
Tamara~G Kolda and Brett~W Bader.
\newblock Tensor decompositions and applications.
\newblock \emph{SIAM Rev.}, 51\penalty0 (3):\penalty0 455--500, 2009.

\bibitem[Kossaifi(2021)]{tensorly-torch}
Jean Kossaifi.
\newblock Tensorly-torch.
\newblock \url{https://github.com/tensorly/torch}, 2021.

\bibitem[Kossaifi et~al.(2019)Kossaifi, Panagakis, Anandkumar, and
  Pantic]{tensorly}
Jean Kossaifi, Yannis Panagakis, Anima Anandkumar, and Maja Pantic.
\newblock Tensorly: Tensor learning in python.
\newblock \emph{Journal of Machine Learning Research (JMLR)}, 20\penalty0 (26),
  2019.

\bibitem[Kossaifi et~al.(2020)Kossaifi, Toisoul, Bulat, Panagakis, Hospedales,
  and Pantic]{kossaifi2020factorized}
Jean Kossaifi, Antoine Toisoul, Adrian Bulat, Yannis Panagakis, Timothy~M
  Hospedales, and Maja Pantic.
\newblock Factorized higher-order {CNN}s with an application to spatio-temporal
  emotion estimation.
\newblock pp.\  6060--6069, 2020.

\bibitem[Kovachki et~al.(2021{\natexlab{a}})Kovachki, Lanthaler, and
  Mishra]{kovachki2021onuniversal}
Nikola Kovachki, Samuel Lanthaler, and Siddhartha Mishra.
\newblock On universal approximation and error bounds for fourier neural
  operators.
\newblock \emph{Journal of Machine Learning Research}, 22\penalty0
  (290):\penalty0 1--76, 2021{\natexlab{a}}.

\bibitem[Kovachki et~al.(2021{\natexlab{b}})Kovachki, Li, Liu, Azizzadenesheli,
  Bhattacharya, Stuart, and Anandkumar]{universal}
Nikola Kovachki, Zongyi Li, Burigede Liu, Kamyar Azizzadenesheli, Kaushik
  Bhattacharya, Andrew Stuart, and Anima Anandkumar.
\newblock Neural operator: Learning maps between function spaces.
\newblock \emph{arXiv preprint arXiv:2108.08481}, 2021{\natexlab{b}}.

\bibitem[Lebedev et~al.(2015)Lebedev, Ganin, Rakhuba, Oseledets, and
  Lempitsky]{lebedev2015speeding}
Vadim Lebedev, Yaroslav Ganin, Maksim Rakhuba, Ivan~V. Oseledets, and Victor~S.
  Lempitsky.
\newblock Speeding-up convolutional neural networks using fine-tuned
  {CP}-decomposition.
\newblock 2015.

\bibitem[Leutbecher \& Palmer(2008)Leutbecher and
  Palmer]{leutbecher2008ensemble}
Martin Leutbecher and Tim~N Palmer.
\newblock Ensemble forecasting.
\newblock \emph{Journal of computational physics}, 227\penalty0 (7):\penalty0
  3515--3539, 2008.

\bibitem[Li et~al.(2020{\natexlab{a}})Li, Kovachki, Azizzadenesheli, Liu,
  Bhattacharya, Stuart, and Anandkumar]{Graph}
Zongyi Li, Nikola Kovachki, Kamyar Azizzadenesheli, Burigede Liu, Kaushik
  Bhattacharya, Andrew Stuart, and Anima Anandkumar.
\newblock Neural operator: Graph kernel network for partial differential
  equations.
\newblock \emph{arXiv preprint arXiv:2003.03485}, 2020{\natexlab{a}}.

\bibitem[Li et~al.(2020{\natexlab{b}})Li, Kovachki, Azizzadenesheli, Liu,
  Stuart, Bhattacharya, and Anandkumar]{Multipole}
Zongyi Li, Nikola Kovachki, Kamyar Azizzadenesheli, Burigede Liu, Andrew
  Stuart, Kaushik Bhattacharya, and Anima Anandkumar.
\newblock Multipole graph neural operator for parametric partial differential
  equations.
\newblock \emph{Advances in Neural Information Processing Systems},
  33:\penalty0 6755--6766, 2020{\natexlab{b}}.

\bibitem[Li et~al.(2021{\natexlab{a}})Li, Kovachki, Azizzadenesheli, Liu,
  Bhattacharya, Stuart, and Anandkumar]{li2021markov}
Zongyi Li, Nikola Kovachki, Kamyar Azizzadenesheli, Burigede Liu, Kaushik
  Bhattacharya, Andrew Stuart, and Anima Anandkumar.
\newblock Markov neural operators for learning chaotic systems.
\newblock \emph{arXiv preprint arXiv:2106.06898}, 2021{\natexlab{a}}.

\bibitem[Li et~al.(2021{\natexlab{b}})Li, Kovachki, Azizzadenesheli, liu,
  Bhattacharya, Stuart, and Anandkumar]{FNO}
Zongyi Li, Nikola~Borislavov Kovachki, Kamyar Azizzadenesheli, Burigede liu,
  Kaushik Bhattacharya, Andrew Stuart, and Anima Anandkumar.
\newblock Fourier neural operator for parametric partial differential
  equations.
\newblock In \emph{International Conference on Learning Representations},
  2021{\natexlab{b}}.

\bibitem[Li et~al.(2021{\natexlab{c}})Li, Zheng, Kovachki, Jin, Chen, Liu,
  Azizzadenesheli, and Anandkumar]{PINO}
Zongyi Li, Hongkai Zheng, Nikola Kovachki, David Jin, Haoxuan Chen, Burigede
  Liu, Kamyar Azizzadenesheli, and Anima Anandkumar.
\newblock Physics-informed neural operator for learning partial differential
  equations.
\newblock \emph{arXiv preprint arXiv:2111.03794}, 2021{\natexlab{c}}.

\bibitem[Liu et~al.(2022)Liu, Kovachki, Li, Azizzadenesheli, Anandkumar,
  Stuart, and Bhattacharya]{liu2022learning}
Burigede Liu, Nikola Kovachki, Zongyi Li, Kamyar Azizzadenesheli, Anima
  Anandkumar, Andrew~M Stuart, and Kaushik Bhattacharya.
\newblock A learning-based multiscale method and its application to inelastic
  impact problems.
\newblock \emph{Journal of the Mechanics and Physics of Solids}, 158:\penalty0
  104668, 2022.

\bibitem[Lu et~al.(2019)Lu, Jin, and Karniadakis]{lu2019deeponet}
Lu~Lu, Pengzhan Jin, and George~Em Karniadakis.
\newblock Deeponet: Learning nonlinear operators for identifying differential
  equations based on the universal approximation theorem of operators.
\newblock \emph{arXiv preprint arXiv:1910.03193}, 2019.

\bibitem[McCormick(1985)]{McCormick1985Multigrid}
S.~F. McCormick.
\newblock Multigrid methods for variational problems: General theory for the v-
  cycle.
\newblock \emph{SIAM Journal on Numerical Analysis}, 22\penalty0 (4):\penalty0
  634--643, 1985.

\bibitem[Novikov et~al.(2015)Novikov, Podoprikhin, Osokin, and
  Vetrov]{novikov2015tensorizing}
Alexander Novikov, Dmitry Podoprikhin, Anton Osokin, and Dmitry Vetrov.
\newblock Tensorizing neural networks.
\newblock pp.\  442--450, 2015.

\bibitem[Oseledets(2011)]{oseledets2011tensor}
I.~V. Oseledets.
\newblock Tensor-train decomposition.
\newblock \emph{SIAM J. Sci. Comput.}, 33\penalty0 (5):\penalty0 2295--2317,
  September 2011.

\bibitem[Panagakis et~al.(2021)Panagakis, Kossaifi, Chrysos, Oldfield,
  Nicolaou, Anandkumar, and Zafeiriou]{tensor2021panagakis}
Yannis Panagakis, Jean Kossaifi, Grigorios~G. Chrysos, James Oldfield,
  Mihalis~A. Nicolaou, Anima Anandkumar, and Stefanos Zafeiriou.
\newblock Tensor methods in computer vision and deep learning.
\newblock \emph{Proceedings of the IEEE}, 109\penalty0 (5):\penalty0 863--890,
  2021.
\newblock \doi{10.1109/JPROC.2021.3074329}.

\bibitem[Papadopoulos et~al.(2022)Papadopoulos, Panagakis, Koubarakis, and
  Nicolaou]{papadopoulos2022efficient}
Christos Papadopoulos, Yannis Panagakis, Manolis Koubarakis, and Mihalis
  Nicolaou.
\newblock Efficient learning of multiple nlp tasks via collective weight
  factorization on bert.
\newblock In \emph{Findings of the Association for Computational Linguistics:
  NAACL 2022}, pp.\  882--890, 2022.

\bibitem[Papalexakis et~al.(2016)Papalexakis, Faloutsos, and
  Sidiropoulos]{papalexakis2016tensors}
Evangelos~E Papalexakis, Christos Faloutsos, and Nicholas~D Sidiropoulos.
\newblock Tensors for data mining and data fusion: Models, applications, and
  scalable algorithms.
\newblock \emph{{ACM} Trans. Intell. Syst. and Technol. (TIST)}, 8\penalty0
  (2):\penalty0 1--44, 2016.

\bibitem[Paszke et~al.(2017)Paszke, Gross, Chintala, Chanan, Yang, DeVito, Lin,
  Desmaison, Antiga, and Lerer]{pytorch}
Adam Paszke, Sam Gross, Soumith Chintala, Gregory Chanan, Edward Yang, Zachary
  DeVito, Zeming Lin, Alban Desmaison, Luca Antiga, and Adam Lerer.
\newblock Automatic differentiation in {P}y{T}orch.
\newblock 2017.

\bibitem[Pathak et~al.(2022)Pathak, Subramanian, Harrington, Raja,
  Chattopadhyay, Mardani, Kurth, Hall, Li, Azizzadenesheli,
  et~al.]{pathak2022fourcastnet}
Jaideep Pathak, Shashank Subramanian, Peter Harrington, Sanjeev Raja, Ashesh
  Chattopadhyay, Morteza Mardani, Thorsten Kurth, David Hall, Zongyi Li, Kamyar
  Azizzadenesheli, et~al.
\newblock Fourcastnet: A global data-driven high-resolution weather model using
  adaptive fourier neural operators.
\newblock \emph{arXiv preprint arXiv:2202.11214}, 2022.

\bibitem[Powell et~al.(2014)Powell, Lord, and
  Shardlow]{powell2014AnIntorduction}
{Catherine E.} Powell, Gabriel Lord, and Tony Shardlow.
\newblock \emph{An Introduction to Computational Stochastic PDEs}.
\newblock Texts in Applied Mathematics. Cambridge University Press, United
  Kingdom, 1 edition, August 2014.
\newblock ISBN 9780521728522.

\bibitem[Rahman et~al.(2022{\natexlab{a}})Rahman, Florez, Anandkumar, Ross, and
  Azizzadenesheli]{rahman2022generative}
Md~Ashiqur Rahman, Manuel~A Florez, Anima Anandkumar, Zachary~E Ross, and
  Kamyar Azizzadenesheli.
\newblock Generative adversarial neural operators.
\newblock \emph{arXiv preprint arXiv:2205.03017}, 2022{\natexlab{a}}.

\bibitem[Rahman et~al.(2022{\natexlab{b}})Rahman, Ross, and
  Azizzadenesheli]{rahman2022u}
Md~Ashiqur Rahman, Zachary~E Ross, and Kamyar Azizzadenesheli.
\newblock U-no: U-shaped neural operators.
\newblock \emph{arXiv preprint arXiv:2204.11127}, 2022{\natexlab{b}}.

\bibitem[Sidiropoulos et~al.(2017)Sidiropoulos, De~Lathauwer, Fu, Huang,
  Papalexakis, and Faloutsos]{sidiropoulos2017tensor}
Nicholas~D Sidiropoulos, Lieven De~Lathauwer, Xiao Fu, Kejun Huang, Evangelos~E
  Papalexakis, and Christos Faloutsos.
\newblock Tensor decomposition for signal processing and machine learning.
\newblock \emph{Transactions Signal Processing}, 65\penalty0 (13):\penalty0
  3551--3582, 2017.

\bibitem[Slingo \& Palmer(2011)Slingo and Palmer]{slingo2011uncertainty}
Julia Slingo and Tim Palmer.
\newblock Uncertainty in weather and climate prediction.
\newblock \emph{Philosophical Transactions of the Royal Society A:
  Mathematical, Physical and Engineering Sciences}, 369\penalty0
  (1956):\penalty0 4751--4767, 2011.

\bibitem[Temam(1988)]{temam1988infinite}
Roger Temam.
\newblock \emph{Infinite-dimensional dynamical systems in mechanics and
  physics}.
\newblock Applied mathematical sciences. Springer-Verlag, New York, 1988.

\bibitem[Tran et~al.(2023)Tran, Mathews, Xie, and Ong]{tran2023factorized}
Alasdair Tran, Alexander Mathews, Lexing Xie, and Cheng~Soon Ong.
\newblock Factorized fourier neural operators.
\newblock In \emph{The Eleventh International Conference on Learning
  Representations}, 2023.
\newblock URL \url{https://openreview.net/forum?id=tmIiMPl4IPa}.

\bibitem[Ulyanov et~al.(2016)Ulyanov, Vedaldi, and Lempitsky]{instancenrom}
Dmitry Ulyanov, Andrea Vedaldi, and Victor Lempitsky.
\newblock Instance normalization: The missing ingredient for fast stylization.
\newblock \emph{arXiv preprint arXiv:1607.08022}, 2016.

\bibitem[Wen et~al.(2022)Wen, Li, Azizzadenesheli, Anandkumar, and Benson]{CCS}
Gege Wen, Zongyi Li, Kamyar Azizzadenesheli, Anima Anandkumar, and Sally~M
  Benson.
\newblock U-fno—an enhanced fourier neural operator-based deep-learning model
  for multiphase flow.
\newblock \emph{Advances in Water Resources}, 163:\penalty0 104180, 2022.

\bibitem[Yang et~al.(2021)Yang, Gao, Castellanos, Ross, Azizzadenesheli, and
  Clayton]{yang2021seismic}
Yan Yang, Angela~F Gao, Jorge~C Castellanos, Zachary~E Ross, Kamyar
  Azizzadenesheli, and Robert~W Clayton.
\newblock Seismic wave propagation and inversion with neural operators.
\newblock \emph{The Seismic Record}, 1\penalty0 (3):\penalty0 126--134, 2021.

\bibitem[Yang et~al.(2022)Yang, Gao, Castellanos, Ross, Azizzadenesheli, and
  Clayton]{yang2022inversion}
Yan Yang, Angela~F Gao, Jorge~C Castellanos, Zachary~E Ross, Kamyar
  Azizzadenesheli, and Robert~W Clayton.
\newblock Accelerated full seismic waveform modeling and inversion with
  u-shaped neural operators.
\newblock \emph{arXiv preprint arXiv:2209.11955}, 2022.

\bibitem[Zhu \& Zabaras(2018)Zhu and Zabaras]{Zabaras}
Yinhao Zhu and Nicholas Zabaras.
\newblock Bayesian deep convolutional encoder–decoder networks for surrogate
  modeling and uncertainty quantification.
\newblock \emph{Journal of Computational Physics}, 2018.
\newblock ISSN 0021-9991.

\end{thebibliography}
